\documentclass[sn-mathphys-num]{sn-jnl}

\usepackage{amsmath,amssymb,amsfonts}%
\usepackage{amsthm}%
\usepackage{graphicx}%
\usepackage{multirow}
\usepackage{booktabs}
\usepackage{xcolor}
\usepackage{textcomp}
\usepackage{manyfoot}
\usepackage[numbers,sort&compress]{natbib}
\usepackage{algorithm}
\usepackage{algpseudocode}
\usepackage{float}
\usepackage{lmodern}

\renewcommand{\orcid}[1]{}
\makeatletter
\@ifundefined{jyear}{\newcommand{\jyear}[1]{}}{}
\makeatother

\jyear{2026}

\title[BFORE: Butterfly-Firefly Optimized Retinex Enhancement]{BFORE: Butterfly-Firefly Optimized Retinex Enhancement for Low-Light Image Quality Improvement}

\author*[1]{\fnm{Ahmed} \sur{Cherif}}\email{ahmed1.cherif@sofrecom.com}

\affil[1]{\orgdiv{Sofrecom Tunisia}, \orgname{Orange Innovation}, \city{Tunis}, \postcode{1053}, \country{Tunisia}}

\begin{document}

\abstract{%
Low-light images suffer from poor visibility, noise, and color distortion. Existing Retinex-based enhancement methods rely on manually tuned parameters that do not generalize across different lighting conditions.

This paper proposes BFORE (Butterfly-Firefly Optimized Retinex Enhancement), a framework that automatically finds the best enhancement parameters for each image. BFORE works in two phases: (1)~a Butterfly Optimization Algorithm (BOA) searches for optimal Multi-Scale Retinex with Color Restoration (MSRCR) parameters, then (2)~a Firefly Algorithm (FA) fine-tunes gamma correction, denoising, and color parameters. Both phases maximize a Gaussian Naturalness Score (GNS)---a no-reference metric that measures how natural the enhanced image looks. Standard quality metrics (PSNR, SSIM, NIQE) are computed only after optimization, ensuring zero data leakage.

On 30 synthetic image pairs, BFORE achieves GNS~=~0.971, outperforming the next-best method MSRCR (0.894) by 8.6\%. On 115 real images from the LOL dataset, BFORE achieves GNS~=~0.887, outperforming MSRCR (0.808) by 9.8\%. A controlled comparison with three deep learning baselines (Zero-DCE, SCI, IAT) trained under identical conditions shows BFORE surpasses the best DL method by 14.7\% in GNS. An ablation study confirms that the hybrid BOA+FA strategy significantly outperforms each optimizer in isolation, and a scalability analysis at three evaluation budgets shows that the structured optimizer significantly outperforms uniform random sampling once compute is available ($p = 0.009$ at 128 evaluations, $p = 0.021$ at 300 evaluations). All improvements are statistically significant ($p < 0.0001$, Wilcoxon signed-rank test). Processing time is 3--6~minutes per image on CPU, suitable for offline applications.
}

\renewcommand{\and}{\unskip{} \textperiodcentered{} }

\maketitle

\section{Introduction}
\label{sec:introduction}

Images captured in low light---at night, in underground environments, or in poorly lit rooms---suffer from low visibility, weak contrast, color distortion, and high noise \cite{loh2019getting,yan2026image}. These degradations reduce both human visual perception and the accuracy of downstream computer vision tasks such as object detection and segmentation \cite{chen2018learning,liu2021benchmarking}.

Low-light image enhancement aims to restore brightness and contrast while preserving natural colors and suppressing noise \cite{land1977retinex}. This is challenging because these objectives often conflict: increasing brightness amplifies noise, and boosting contrast can distort colors \cite{abdullah2007dynamic,ibrahim2007brightness}.

Existing enhancement methods fall into three categories. \textbf{Histogram-based methods} such as HE and CLAHE \cite{gonzalez2018digital,zuiderveld1994contrast} redistribute pixel intensities to improve contrast, but often produce artifacts and unnatural results. \textbf{Retinex-based methods} \cite{land1971lightness,jobson1997properties} decompose images into illumination and reflectance components, offering a more principled approach; however, their quality depends heavily on manually chosen parameters (Gaussian scales, weights, color restoration coefficients) that remain fixed across images \cite{petro2014multiscale,provenzi2005mathematical}. \textbf{Deep learning methods} such as RetinexNet \cite{wei2018deep}, KinD \cite{zhang2019kindling}, EnlightenGAN \cite{jiang2021enlightengan}, Zero-DCE \cite{guo2020zero}, and MIRNet \cite{zamir2020learning} achieve strong results but require large training datasets, significant computational resources, and may not generalize to unseen domains \cite{liu2021retinex,xu2023lowlight}.

The core limitation of Retinex-based methods is their reliance on fixed parameters. The Gaussian scales ($\sigma_1, \sigma_2, \sigma_3$), weights ($\omega_1, \omega_2, \omega_3$), color restoration gain ($\mu$), and post-processing parameters (gamma, denoising threshold) are typically set once and used for all images \cite{ma2017multiscale,sadia2019color}. This one-size-fits-all approach produces suboptimal results because different images require different settings. While some works have explored adaptive strategies \cite{ning2023low,parthasarathy2012automated}, they address only individual stages rather than the full pipeline.

\textbf{The gap:} No existing work has used hybrid metaheuristic optimization to simultaneously tune all parameters of a multi-stage Retinex pipeline---including MSRCR, gamma correction, and denoising---in a single integrated framework.

To address this gap, we propose \textbf{BFORE} (Butterfly-Firefly Optimized Retinex Enhancement). BFORE combines two nature-inspired optimization algorithms:
\begin{itemize}
    \item \textbf{Butterfly Optimization Algorithm (BOA)} \cite{arora2019butterfly}---inspired by butterfly foraging behavior, BOA performs broad global search to find promising parameter regions.
    \item \textbf{Firefly Algorithm (FA)} \cite{yang2009firefly}---inspired by firefly light attraction, FA performs precise local refinement within promising regions.
\end{itemize}
By running BOA first (global search) then FA (local fine-tuning), BFORE automatically finds the best enhancement parameters for each individual image.

The main contributions are:

\begin{enumerate}
    \item \textbf{Automatic parameter tuning:} BFORE eliminates manual parameter selection by automatically optimizing all 16 parameters of a multi-stage Retinex pipeline for each input image.
    \item \textbf{Hybrid BOA-FA optimization:} A two-phase strategy where BOA explores the MSRCR parameter space globally and FA refines LAGC/ANLM parameters locally, with convergence-based phase switching.
    \item \textbf{Integrated enhancement pipeline:} A complete pipeline combining gamma correction, denoising, color correction, saturation enhancement, and MSRCR color restoration---all jointly optimized via a no-reference GNS fitness function.
    \item \textbf{Comprehensive evaluation:} Testing on synthetic ($n=30$) and real LOL ($n=115$) datasets, controlled comparison with three deep learning baselines (Zero-DCE, SCI, IAT) under identical conditions, a four-variant ablation study, and a scalability analysis at three evaluation budgets showing that the structured BOA+FA optimizer significantly outperforms uniform random sampling once compute is available. All GNS improvements over classical baselines are statistically significant ($p < 0.0001$, Wilcoxon signed-rank test).
\end{enumerate}

The rest of this paper is organized as follows. Section~\ref{sec:related} reviews related work on low-light image enhancement, Retinex-based methods, and metaheuristic optimization algorithms. Section~\ref{sec:method} presents the proposed BFORE framework in detail, including the multi-stage enhancement pipeline and the hybrid BOA-FA optimization strategy. Section~\ref{sec:experiments} describes the experimental setup, benchmark datasets, evaluation metrics, and presents comprehensive results including comparisons with state-of-the-art methods and ablation studies. Section~\ref{sec:conclusion} concludes the paper and outlines directions for future work.

\section{Related Work}
\label{sec:related}

We review three areas relevant to BFORE: traditional enhancement methods, deep learning approaches, and metaheuristic optimization for image processing.

\subsection{Traditional image enhancement methods}
\label{sec:traditional}

Traditional low-light image enhancement methods can be broadly classified into histogram-based, Retinex-based, and transform-domain approaches.

\textbf{Histogram-based methods.} Histogram Equalization (HE) \cite{gonzalez2018digital} improves contrast by flattening the intensity histogram, but often produces over-enhanced, unnatural results. CLAHE \cite{zuiderveld1994contrast} mitigates this by applying local equalization with a contrast limit. BBHE \cite{kim1997contrast} preserves mean brightness by splitting the histogram. Despite these improvements, histogram-based methods lack a physical model of image formation and cannot distinguish illumination from reflectance.

\textbf{Retinex-based methods.} Retinex theory \cite{land1971lightness} models a captured image as $S = L \cdot R$, where $L$ is illumination and $R$ is reflectance. Enhancement recovers $R$ by estimating and removing $L$. Jobson et al.\ \cite{jobson1997multiscale} introduced Multi-Scale Retinex (MSR), combining Gaussian surround functions at different scales, and later added color restoration (MSRCR) \cite{jobson1997properties}. However, MSRCR quality depends heavily on Gaussian scales ($\sigma_1, \sigma_2, \sigma_3$), weights, and color restoration coefficients, which are typically fixed \cite{petro2014multiscale}. Subsequent work addressed specific limitations: Li et al.\ \cite{li2020enhanced} used improved bootstrap filtering for mine images; Su et al.\ \cite{su2020multiscale} employed gradient domain-guided filtering; Zhi et al.\ \cite{zhi2022enhanced} combined illuminance adjustment with CLAHE; Zhang et al.\ \cite{zhang2021improved} fused bilateral filtering with multi-scale Retinex. All these methods improve upon basic Retinex but still rely on empirically determined parameters that do not adapt to individual images.

\textbf{Other traditional methods.} Adaptive Gamma Correction with Weighted Distribution (AGCWD) \cite{huang2013efficient} dynamically adjusts the gamma value based on a CDF-weighted histogram, producing a global tone-mapping curve. However, standard AGCWD may amplify noise in extremely dark regions and lacks spatial adaptivity. Guided Image Filtering \cite{he2013guided} provides edge-preserving smoothing that is useful for post-processing enhanced images. The BM3D algorithm \cite{dabov2007image} offers powerful denoising through block-matching and 3D collaborative filtering, while Non-Local Means (NLM) \cite{buades2005non} denoises by weighted averaging of similar patches across the image, providing effective noise suppression with good detail preservation at lower computational cost than BM3D.

\subsection{Deep learning-based enhancement methods}
\label{sec:deeplearning}

The advent of deep learning has brought significant advances to low-light image enhancement. Wei et al.\ \cite{wei2018deep} proposed RetinexNet, which combines Retinex theory with deep neural networks by learning the decomposition of images into reflectance and illumination maps. Zhang et al.\ \cite{zhang2019kindling} introduced KinD (Kindling the Darkness), a two-branch network for layer decomposition and adjustment with degradation-aware training. Jiang et al.\ \cite{jiang2021enlightengan} proposed EnlightenGAN, an unsupervised generative adversarial network that does not require paired training data. Guo et al.\ \cite{guo2020zero} introduced Zero-DCE, which formulates light enhancement as a task of image-specific curve estimation, enabling training without paired or unpaired data. Zamir et al.\ \cite{zamir2020learning} proposed MIRNet, a multi-scale residual block architecture that learns an enriched set of features across multiple spatial resolutions.

More recent works explore transformer-based architectures \cite{wang2023ultra}, initialization-aware Retinex decomposition \cite{fan2025iniretinex}, and state-space models \cite{yin2025mambadpf}. While these methods achieve strong results, they require large training datasets and significant compute, and may not generalize to unseen domains \cite{liu2021retinex,xu2023lowlight}. Their black-box nature also limits interpretability in safety-critical applications.

\subsection{Metaheuristic optimization for image processing}
\label{sec:metaheuristic}

Metaheuristic optimization algorithms have gained significant attention for solving complex optimization problems in image processing due to their ability to explore large, non-convex search spaces without requiring gradient information \cite{mirjalili2019genetic,kennedy1995particle}.

\textbf{Butterfly Optimization Algorithm (BOA).} Proposed by Arora and Singh \cite{arora2019improved}, BOA is inspired by butterfly foraging behavior, using fragrance-based attraction with global and local search phases controlled by a switch probability. BOA has been applied to various optimization problems \cite{makhadmeh2023recent,sharma2022butterfly}, but not to image enhancement parameter optimization.

\textbf{Firefly Algorithm (FA).} Proposed by Yang \cite{yang2009firefly}, FA models firefly attraction where brighter fireflies attract dimmer ones, with attractiveness decreasing with distance. FA has been applied to image segmentation \cite{horng2012vector}, feature selection \cite{arora2018novel}, and image filtering \cite{tilahun2012modified}.

\textbf{Hybrid metaheuristic approaches.} Hybridizing complementary search mechanisms---global exploration with local exploitation---improves optimization performance \cite{talbi2002taxonomy,fister2013modified}. In image processing, hybrid approaches have been applied to segmentation \cite{bhandari2015modified} and clustering \cite{jyothi2026robust}. For low-light enhancement, Deng et al.\ \cite{deng2022low} combined Artificial Bee Colony with Multi-Scale Retinex, demonstrating the potential of bio-inspired optimization for Retinex parameter tuning. However, no existing work uses a hybrid BOA-FA strategy to simultaneously tune all parameters of a multi-stage Retinex pipeline including MSRCR, gamma correction, and denoising.

\section{Proposed Method}
\label{sec:method}

This section describes BFORE in detail: the overall architecture, each processing stage, and the hybrid BOA-FA optimization strategy.

\subsection{Overall architecture}
\label{sec:architecture}

\begin{figure}[htbp]
    \centering
    \includegraphics[width=\textwidth]{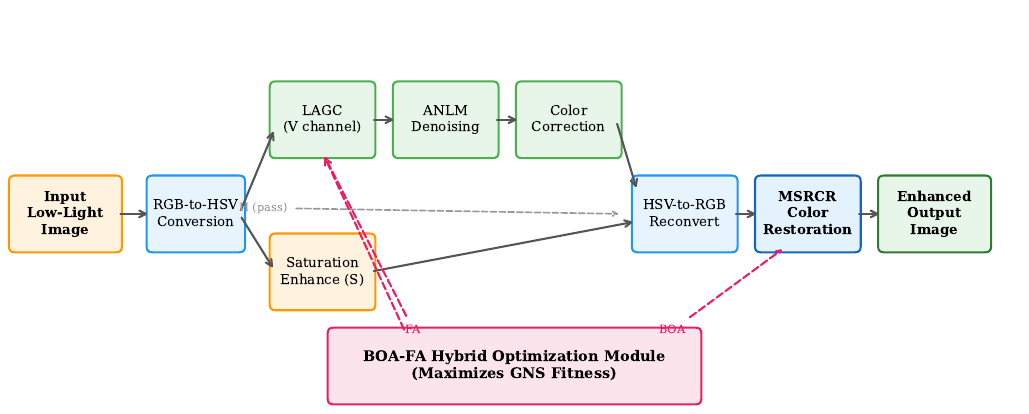}
    \caption{Overall architecture of the proposed BFORE framework. The input low-light image is converted to HSV color space. The V channel undergoes LAGC (FA-optimized) and ANLM denoising, while the S channel is enhanced via contrast stretching. After HSV-to-RGB reconversion, MSRCR color restoration (BOA-optimized) is applied. The BOA-FA optimization module automatically tunes all pipeline parameters by maximizing the GNS fitness.}
    \label{fig:architecture}
\end{figure}

The overall architecture of the proposed BFORE framework is illustrated in Fig.~\ref{fig:architecture}. The framework consists of two major components: (1) a multi-stage Retinex-based enhancement pipeline and (2) a hybrid BOA-FA optimization module that automatically tunes the pipeline parameters.

The enhancement pipeline operates as follows:

\begin{enumerate}
    \item \textbf{Color space conversion}: The input RGB image is converted to HSV color space to separate the luminance (V) channel from the chrominance (H, S) channels, enabling independent processing of brightness information without introducing color distortion.
    \item \textbf{Adaptive luminance correction}: The V channel undergoes Local Adaptive Gamma Correction (LAGC) to improve brightness and contrast, with parameters optimized by the Firefly Algorithm.
    \item \textbf{Adaptive denoising}: The corrected V channel is processed by an Adaptive Non-Local Means (ANLM) denoising algorithm with FA-optimized parameters to suppress noise while preserving edge details.
    \item \textbf{Color correction}: A nonlinear color correction is applied to address chromaticity deviations introduced during luminance processing.
    \item \textbf{Saturation enhancement}: The S channel undergoes contrast stretching to improve color saturation and visual appearance.
    \item \textbf{Color space reconversion}: The processed H, S, and V channels are recombined and converted back to RGB color space.
    \item \textbf{Global color restoration}: The MSRCR algorithm with BOA-optimized parameters is applied to the reconverted RGB image for final global color correction and dynamic range adjustment.
\end{enumerate}

\subsection{Retinex theory and MSRCR formulation}
\label{sec:retinex}

According to Retinex theory \cite{land1971lightness}, the observed image signal $S(x, y)$ captured by a camera can be decomposed into two components: the incident illumination $L(x, y)$ and the reflected light $R(x, y)$:
\begin{equation}
    S(x, y) = L(x, y) \cdot R(x, y)
    \label{eq:retinex}
\end{equation}
where $L(x, y)$ represents the illumination component of the image, and $R(x, y)$ denotes the reflectance component.

The Single Scale Retinex (SSR) algorithm estimates the reflectance by subtracting the estimated illumination (obtained via Gaussian convolution) from the original image in the logarithmic domain:
\begin{equation}
    R_i(x, y) = \log I_i(x, y) - \log[I_i(x, y) * f(x, y)]
    \label{eq:ssr}
\end{equation}
where $I_i(x, y)$ is the $i$-th color band of the original image, $*$ denotes the convolution operation, and $f(x, y)$ is a Gaussian surround function defined as:
\begin{equation}
    f(x, y) = \lambda \cdot e^{-\frac{x^2 + y^2}{2\sigma^2}}
    \label{eq:gaussian}
\end{equation}
where $\sigma$ is the standard deviation (scale parameter) that controls the spatial extent of the surround function, and $\lambda$ is a normalization constant satisfying $\iint f(x, y)\,dx\,dy = 1$.

The Multi-Scale Retinex (MSR) algorithm extends SSR by combining outputs from $N$ different Gaussian scales to achieve a balance between dynamic range compression and tonal rendition:
\begin{equation}
    R_{\text{MSR}}(x, y) = \sum_{k=1}^{N} \omega_k \left\{ \log I_i(x, y) - \log[I_i(x, y) * f_k(x, y)] \right\}
    \label{eq:msr}
\end{equation}
where $N$ is the number of scales (typically $N = 3$), $\omega_k$ is the weight for the $k$-th scale satisfying $\sum \omega_k = 1$, and $f_k(x, y)$ is the Gaussian surround function with scale parameter $\sigma_k$.

The Multi-Scale Retinex with Color Restoration (MSRCR) further incorporates a color recovery factor $C$ to address color desaturation:
\begin{equation}
    R_{\text{MSRCR}}(x, y) = C_j(x, y) \cdot R_{\text{MSR}}(x, y)
    \label{eq:msrcr}
\end{equation}
\begin{equation}
    C_j(x, y) = \mu \cdot \log\left[\eta \cdot \frac{I(x, y)}{\sum_{n=1}^{N_c} I_n(x, y)}\right]
    \label{eq:color_restoration}
\end{equation}
where $j$ indexes the color channel, $\mu$ is the gain coefficient, $\eta$ is the nonlinear intensity factor, and $N_c$ is the number of color channels.

\textbf{Parameters to be optimized by BOA}: The MSRCR algorithm involves the following parameters that significantly affect enhancement quality: $\{\sigma_1, \sigma_2, \sigma_3, \omega_1, \omega_2, \omega_3, \mu, \eta\}$. Traditional implementations use fixed default values (e.g., $\sigma_1 = 15$, $\sigma_2 = 80$, $\sigma_3 = 250$, $\omega_1 = \omega_2 = \omega_3 = 1/3$), which are suboptimal for varying image characteristics. The proposed BFORE framework employs BOA to automatically optimize these parameters for each input image.

\subsection{Local Adaptive Gamma Correction (LAGC)}
\label{sec:lagc}

Standard gamma correction transforms pixel values through a power function:
\begin{equation}
    T(l) = l_{\max}\left(\frac{l}{l_{\max}}\right)^{\gamma}
    \label{eq:gamma}
\end{equation}
where $T(l)$ is the transformed pixel value, $l$ and $l_{\max}$ are the pixel value and maximum pixel value, respectively, and $\gamma$ controls the degree of correction.

The proposed improved LAGC computes a spatially varying gamma value based on local image statistics. For a point $(x, y)$ with a local statistics kernel of size $M \times M$, the local mean $\mu_L$ and local standard deviation $\sigma_L$ are computed as:
\begin{equation}
    \mu_L = \frac{1}{M^2}\sum_{i=1}^{M}\sum_{j=1}^{M} x(i, j)
    \label{eq:local_mean}
\end{equation}
\begin{equation}
    \sigma_L = \sqrt{\frac{1}{M^2}\sum_{i=1}^{M}\sum_{j=1}^{M} (x(i, j) - \mu_L)^2}
    \label{eq:local_std}
\end{equation}

The adaptive gamma value for each pixel is then computed as:
\begin{equation}
    \gamma(x, y) = -\frac{\log_2(\sigma_L + \mu_L)}{2}
    \label{eq:adaptive_gamma}
\end{equation}

To prevent noise amplification in extremely dark, near-homogeneous regions where $\sigma_L$ approaches zero, a noise floor clamping is introduced:
\begin{equation}
    \sigma_L' = \max(\sigma_L, \sigma_{\min})
    \label{eq:noise_floor}
\end{equation}
where $\sigma_{\min}$ is a small positive threshold (empirically set within $[3, 8]$ on a 0--255 scale). The corrected pixel value is then:
\begin{equation}
    F(i, j) = f_{\max}\left(\frac{f(i, j)}{f_{\max}}\right)^{\gamma}
    \label{eq:corrected_pixel}
\end{equation}

\textbf{Parameters to be optimized by FA}: The LAGC stage involves the following parameters: $\{\gamma_0$ (base gamma), $M$ (local statistics kernel size), $\sigma_{\min}$ (noise floor threshold)$\}$. These parameters directly affect the brightness adaptation behavior and noise sensitivity of the enhancement.

\textbf{Distinction from AGCWD \cite{huang2013efficient}.} The original AGCWD by Huang et al.\ computes a single global gamma from the normalized CDF of the intensity histogram, producing a monotonic tone-mapping curve. In contrast, LAGC computes a \emph{spatially varying} gamma from local statistics within an $M \times M$ kernel around each pixel (Eq.~\ref{eq:adaptive_gamma}), enabling pixel-wise brightness adaptation. AGCWD is included as a baseline in Table~\ref{tab:quantitative}.

\subsection{Adaptive Non-Local Means (ANLM) Denoising}
\label{sec:anlm}

The enhancement process may amplify noise present in the original low-light image. To address this, an Adaptive Non-Local Means (ANLM) denoising algorithm is applied to the corrected luminance channel. Non-Local Means (NLM) \cite{buades2005non} denoises by replacing each pixel with a weighted average of similar patches across the image, where similarity is measured in a local search window.

The NLM denoised value for pixel $x$ is:
\begin{equation}
    \hat{v}(x) = \frac{\sum_{y \in \mathcal{N}(x)} w(x,y)\, v(y)}{\sum_{y \in \mathcal{N}(x)} w(x,y)}
    \label{eq:nlm_basic}
\end{equation}
where $\mathcal{N}(x)$ is the search window around pixel $x$, $v(y)$ is the noisy pixel value, and the weight $w(x,y)$ is computed from patch similarity:
\begin{equation}
    w(x,y) = \exp\!\left(-\frac{\|P(x) - P(y)\|_2^2}{h^2}\right)
    \label{eq:nlm_weight}
\end{equation}
where $P(x)$ and $P(y)$ are patches centered at $x$ and $y$, and $h$ is the filtering strength parameter that controls the degree of smoothing.

The key innovation of ANLM over standard NLM is the adaptive selection of $h$ via the FA optimizer: instead of requiring prior knowledge of noise variance, the filtering strength is tuned jointly with the LAGC parameters to maximize the GNS fitness on the current image.

\textbf{ANLM parameters to be optimized by FA}: $\{h$ (filtering strength), block\_size (patch size for matching)$\}$.

\subsection{Color correction and saturation enhancement}
\label{sec:color}

After processing the luminance channel, a nonlinear color correction is applied to address chromaticity deviations:
\begin{equation}
    C_{\text{out}} = L_{\text{out}} \times \left(\frac{C_{\text{in}}}{L_{\text{in}}}\right)^r
    \label{eq:color_correction}
\end{equation}
where $r$ controls the saturation of color correction, $C_{\text{in}}$ represents the R, G, B channels of the input, $L_{\text{in}}$ denotes the luminance value after HSV conversion, $L_{\text{out}}$ indicates the processed luminance, and $C_{\text{out}}$ represents the corrected color channels.

For saturation enhancement, contrast stretching is applied to the S channel in HSV space:
\begin{equation}
    S_{\text{out}} = (S_{\text{in}} - \min) \cdot \frac{N_{\max} - N_{\min}}{\max - \min} + N_{\min}
    \label{eq:saturation}
\end{equation}
where $S_{\text{in}}$ and $S_{\text{out}}$ are input and output saturation values, $\min$ and $\max$ are the original range, and $N_{\min}$ and $N_{\max}$ are the target range.

\subsection{Butterfly Optimization Algorithm (BOA) for MSRCR optimization}
\label{sec:boa}

The Butterfly Optimization Algorithm \cite{arora2019butterfly} is employed to optimize the MSRCR parameters. In BOA, each butterfly in the population represents a candidate solution (a set of MSRCR parameters), and the fragrance emitted by each butterfly is related to the fitness of its solution.

The fragrance perceived by a butterfly is modeled as:
\begin{equation}
    f_i = c \cdot I_i^a
    \label{eq:fragrance}
\end{equation}
where $c$ is the sensory modality ($c \in [0, 1]$), $I_i$ is the stimulus intensity (fitness value) of butterfly $i$, and $a$ is the power exponent ($a \in [0, 1]$) controlling the degree of absorption.

The position update of butterfly $i$ follows either a global search or local search phase, controlled by a switch probability $p$:

\textbf{Global search} (with probability $p$): The butterfly moves toward the current best solution $g^*$:
\begin{equation}
    x_i^{t+1} = x_i^t + (r^2 \cdot g^* - x_i^t) \cdot f_i
    \label{eq:boa_global}
\end{equation}

\textbf{Local search} (with probability $1 - p$): The butterfly moves toward a randomly selected pair of neighbors:
\begin{equation}
    x_i^{t+1} = x_i^t + (r^2 \cdot x_j^t - x_k^t) \cdot f_i
    \label{eq:boa_local}
\end{equation}
where $r$ is a random number in $[0, 1]$, and $x_j$ and $x_k$ are randomly selected butterflies from the population.

In the proposed BFORE framework, each butterfly encodes a parameter vector $\theta_{\text{BOA}} = \{\sigma_1, \sigma_2, \sigma_3, \omega_1, \omega_2, \omega_3, \mu, \eta\}$ with the search bounds specified in Table~\ref{tab:boa_params}.

\begin{table}[htbp]
    \centering
    \caption{BOA search bounds for MSRCR parameters.}
    \label{tab:boa_params}
    \begin{tabular}{lccc}
        \toprule
        \textbf{Parameter} & \textbf{Lower Bound} & \textbf{Upper Bound} & \textbf{Description} \\
        \midrule
        $\sigma_1$ & 5 & 30 & Small-scale Gaussian std \\
        $\sigma_2$ & 50 & 150 & Medium-scale Gaussian std \\
        $\sigma_3$ & 150 & 350 & Large-scale Gaussian std \\
        $\omega_1, \omega_2, \omega_3$ & 0.1 & 0.6 & Scale weights (normalized) \\
        $\mu$ & 80 & 150 & Color restoration gain \\
        $\eta$ & 100 & 200 & Nonlinear intensity \\
        \bottomrule
    \end{tabular}
\end{table}

\subsection{Firefly Algorithm (FA) for LAGC and ANLM optimization}
\label{sec:fa}

The Firefly Algorithm \cite{yang2009firefly} is employed to optimize the LAGC and ANLM parameters. In FA, each firefly represents a candidate solution, and brighter fireflies attract less bright ones.

The light intensity of firefly $i$ at distance $r$ is:
\begin{equation}
    I(r) = I_0 \cdot e^{-\gamma_f r^2}
    \label{eq:light_intensity}
\end{equation}
where $I_0$ is the initial light intensity (fitness value at $r = 0$) and $\gamma_f$ is the light absorption coefficient.

The attractiveness of a firefly is defined as:
\begin{equation}
    \beta(r) = \beta_0 \cdot e^{-\gamma_f r^2}
    \label{eq:attractiveness}
\end{equation}
where $\beta_0$ is the initial attractiveness at $r = 0$.

The movement of firefly $i$ toward a brighter firefly $j$ is:
\begin{equation}
    x_i^{t+1} = x_i^t + \beta_0 \cdot e^{-\gamma_f r_{ij}^2} \cdot (x_j^t - x_i^t) + \alpha \cdot \epsilon_i
    \label{eq:fa_movement}
\end{equation}
where $\alpha$ is the randomization parameter controlling step size, $r_{ij}$ is the Euclidean distance between fireflies $i$ and $j$, and $\epsilon_i$ is a vector of random numbers drawn from a uniform or Gaussian distribution.

Each firefly encodes a parameter vector $\theta_{\text{FA}} = \{\gamma_0, M, \sigma_{\min}, h, \text{block\_size}, \text{sat\_scale}, r\}$ with the search bounds specified in Table~\ref{tab:fa_params}.

\begin{table}[htbp]
    \centering
    \caption{FA search bounds for LAGC and ANLM parameters.}
    \label{tab:fa_params}
    \begin{tabular}{lccc}
        \toprule
        \textbf{Parameter} & \textbf{Lower Bound} & \textbf{Upper Bound} & \textbf{Description} \\
        \midrule
        $\gamma_0$ & 0.3 & 1.5 & Base gamma value \\
        $M$ & 3 & 15 & LAGC local statistics kernel size \\
        $\sigma_{\min}$ & 2 & 10 & Noise floor threshold \\
        $h$ & 5 & 15 & ANLM filtering strength \\
        block\_size & 5 & 11 & ANLM patch size \\
        sat\_scale & 0.5 & 1.5 & Saturation enhancement scale \\
        $r$ & 0.5 & 2.0 & Color correction exponent \\
        \bottomrule
    \end{tabular}
\end{table}

\subsection{Hybrid BOA-FA optimization strategy}
\label{sec:hybrid}

The proposed hybrid BOA-FA strategy leverages the complementary strengths of both algorithms: BOA's strong global exploration capability (through fragrance-based search) and FA's efficient local exploitation capability (through attractiveness-based movement). The switching between the two algorithms is governed by a convergence monitoring mechanism.

\textbf{Fitness function.}
To ensure a data-leakage-free experimental protocol, the optimization fitness function
uses \emph{exclusively no-reference metrics}.
PSNR and SSIM are reserved for the held-out final evaluation and are
never observed by the optimizer.

We define a \emph{Gaussian naturalness score} (GNS) fitness in which each
sub-score is a Gaussian bell centered on the corresponding statistic of
well-exposed natural indoor images (derived from LOL training-set statistics):
\begin{equation}
  \phi(x;\,\mu_0,\sigma_0) \;=\; \exp\!\left(-\frac{(x-\mu_0)^2}{2\sigma_0^2}\right)
  \;\in\;(0,1],
  \label{eq:gaussian_kernel}
\end{equation}
and the overall fitness is:
\begin{equation}
  \mathcal{F}_{\text{NR}}(\theta)
  = 0.30\,\phi\!\bigl(H(\theta);\,7.0,\,1.2\bigr)
  + 0.20\,\phi\!\bigl(\text{AG}(\theta);\,17.0,\,8.0\bigr)
  + 0.10\,\phi\!\bigl(\sigma(\theta);\,40.0,\,18.0\bigr)
  + 0.15\,\phi\!\bigl(\bar{\mu}(\theta);\,128.0,\,35.0\bigr)
  + 0.15\,\phi\!\bigl(\alpha_{\text{MSCN}}(\theta);\,2.0,\,0.8\bigr)
  + 0.10\,\phi\!\bigl(r_{\text{clip}}(\theta);\,0.01,\,0.03\bigr),
  \label{eq:fitness}
\end{equation}
where $H$ is Shannon entropy, $\text{AG}$ is average gradient (sharpness),
$\sigma$ is standard deviation (contrast), $\bar{\mu}$ is mean luminance,
$\alpha_{\text{MSCN}}$ is the generalised Gaussian shape parameter of the
mean-subtracted contrast-normalized (MSCN) coefficients \cite{mittal2012no}
(natural images yield $\alpha\!\approx\!2$), and $r_{\text{clip}}$ is the
fraction of saturated pixels ($\le 5$ or $\ge 250$ on an 8-bit scale).
The two naturalness regularizers prevent the optimizer from improving its
score through over-enhancement artifacts that degrade independent metrics.
The target centres $(\mu_0)$ and widths $(\sigma_0)$ are calibrated from natural well-exposed indoor image statistics, consistent with published naturalness priors \cite{wang2013naturalness,mittal2013making}
(see Table~\ref{tab:fitness_targets}).

\begin{table}[h]
\centering
\caption{GNS fitness target statistics calibrated from natural well-exposed image statistics \cite{wang2013naturalness,mittal2012no}.}
\label{tab:fitness_targets}
\begin{tabular}{lccc}
\toprule
Metric & Target $\mu_0$ & Width $\sigma_0$ & Weight \\
\midrule
Entropy $H$ (bits)    & 7.0  & 1.2  & 0.30 \\
Average Gradient AG   & 17.0 & 8.0 & 0.20 \\
Std.\ deviation $\sigma$ & 40.0 & 18.0 & 0.10 \\
Mean luminance $\bar{\mu}$ & 128.0 & 35.0 & 0.15 \\
MSCN shape $\alpha_{\text{MSCN}}$ & 2.0 & 0.8 & 0.15 \\
Clipping ratio $r_{\text{clip}}$ & 0.01 & 0.03 & 0.10 \\
\bottomrule
\end{tabular}
\end{table}

Because every sub-score is bounded in $(0,1]$ and the Gaussian kernel is
smooth and differentiable everywhere, the fitness landscape is well-conditioned
across all image types---including very dark inputs where a quadratic penalty
term would otherwise dominate and collapse the search to a single dimension.
After optimization converges, PSNR and SSIM are computed once on the
\emph{held-out} enhanced image against the ground-truth reference
(Section~\ref{sec:metrics}), completing a strict two-phase protocol with
zero ground-truth leakage.

\textbf{Convergence-based switching.} BFORE uses a two-phase sequential strategy:
Phase~1 (BOA) optimizes MSRCR and blend parameters; Phase~2 (FA) refines
LAGC and denoising parameters using the MSRCR configuration fixed from Phase~1.
This separation exploits the complementary sensitivity of each algorithm:
BOA's fragrance-based global search is effective for the continuous MSRCR scales
($\sigma_1,\sigma_2,\sigma_3$) and blend ratio $\alpha$, while FA's
normalized-distance attractiveness is more stable for the discrete/bounded LAGC
parameters (base gamma, kernel size).

\begin{algorithm}[htbp]
\caption{BFORE Two-Phase Hybrid BOA-FA Optimization}
\label{alg:bfore}
\begin{algorithmic}
\Require Low-light image $I_{\text{input}}$
\Ensure Enhanced image $I_{\text{enhanced}}$ with optimized parameters $\theta^*$
\State Convert $I_{\text{input}}$ from BGR to HSV $\rightarrow$ $(H, S, V)$

\textbf{--- Phase 1: BOA optimizes MSRCR parameters ---}
\State Initialize BOA population $P = \{\theta_1^{\mathrm{BOA}},\ldots,\theta_{N_B}^{\mathrm{BOA}}\}$ randomly in $\Omega_{\mathrm{BOA}}$
\State Evaluate $f_i = \mathcal{F}_{\mathrm{NR}}(\mathrm{BFORE}(I; \theta_i^{\mathrm{BOA}}, \theta_{\mathrm{def}}^{\mathrm{FA}}))$ for all $i$
\State $\theta^* \leftarrow \arg\max_i f_i$
\For{$t = 1, \ldots, T_{\mathrm{BOA}}$}
    \State $c_t \leftarrow 0.01 + \frac{t-1}{T_{\mathrm{BOA}}-1} \cdot 0.04$ \Comment{Linear c schedule}
    \For{each butterfly $i$}
        \State $f_i \leftarrow c_t \cdot (\mathrm{fitness}_i)^a$
        \If{$r \sim U[0,1] < p$}
            \State $\theta_i \leftarrow \theta_i + r_2^2 \cdot (\theta^* - \theta_i) \cdot f_i$ \Comment{Global search}
        \Else
            \State $\theta_i \leftarrow \theta_i + r_2^2 \cdot (\theta_j - \theta_k) \cdot f_i$, $j,k$ random \Comment{Local search}
        \EndIf
        \If{$r_3 \sim U[0,1] < 0.15$}
            \State $\theta_i \leftarrow \theta_i + 0.01 \cdot L(\beta{=}1.5) \odot s$ \Comment{Lévy flight, $s$ = range scales}
        \EndIf
        \State Clamp $\theta_i$ to $\Omega_{\mathrm{BOA}}$; evaluate $\mathcal{F}_{\mathrm{NR}}(\theta_i)$; update $\theta^*$
    \EndFor
\EndFor
\State $\theta^*_{\mathrm{BOA}} \leftarrow \theta^*$ \Comment{Fix MSRCR parameters}

\textbf{--- Phase 2: FA refines LAGC/ANLM parameters ---}
\State Initialize FA population $P = \{\phi_1^{\mathrm{FA}},\ldots,\phi_{N_F}^{\mathrm{FA}}\}$ in $\Omega_{\mathrm{FA}}$; seed $\phi_1 \leftarrow \phi_{\mathrm{def}}$ (warm-start)
\State Evaluate $f_i = \mathcal{F}_{\mathrm{NR}}(\mathrm{BFORE}(I; \theta^*_{\mathrm{BOA}}, \phi_i))$ for all $i$
\State $\phi^* \leftarrow \arg\max_i f_i$
\For{$t = 1, \ldots, T_{\mathrm{FA}}$}
    \State $\alpha_t \leftarrow \alpha_0 \cdot 0.9^{t-1}$ \Comment{Decaying random walk}
    \For{each firefly $i$}
        \For{each firefly $j$ where $f_j > f_i$}
            \State $\hat{r}_{ij} \leftarrow \|(\phi_i - \phi_j) \oslash s\|_2$ \Comment{Normalised distance}
            \State $\beta \leftarrow \beta_0 e^{-\gamma \hat{r}_{ij}^2}$ \Comment{Attractiveness}
            \State $\phi_i \leftarrow \phi_i + \beta(\phi_j-\phi_i) + \alpha_t \cdot s \odot (U[0,1]^d - 0.5)$
            \State Clamp $\phi_i$ to $\Omega_{\mathrm{FA}}$
        \EndFor
        \State Evaluate $\mathcal{F}_{\mathrm{NR}}(\phi_i)$; update $\phi^*$
    \EndFor
\EndFor
\State $\theta^*_{\mathrm{FA}} \leftarrow \phi^*$

\State Apply $\theta^*_{\mathrm{BOA}}$, $\theta^*_{\mathrm{FA}}$ to full BFORE pipeline on $I_{\text{input}}$
\State Compute PSNR, SSIM on $I_{\text{enhanced}}$ vs.\ ground truth (post-hoc, no leakage)
\State \Return $I_{\text{enhanced}}$
\end{algorithmic}
\end{algorithm}

\subsection{Computational complexity analysis}
\label{sec:complexity}

The computational complexity of the proposed BFORE framework is analyzed as follows. Let $n$ be the total number of pixels in the input image, $N_B$ be the BOA population size, $N_F$ be the FA population size, and $T$ be the total number of optimization iterations.

\begin{itemize}
    \item \textbf{HSV conversion}: $O(n)$
    \item \textbf{LAGC}: $O(n \cdot M^2)$ where $M$ is the local statistics kernel size
    \item \textbf{ANLM denoising}: $O(n \cdot W^2 \cdot B^2)$ where $W$ is the search window size and $B$ is the patch size
    \item \textbf{MSRCR}: $O(n \cdot N \cdot \log n)$ where $N$ is the number of scales (FFT-based convolution)
    \item \textbf{BOA optimization}: $O(T_{\text{BOA}} \cdot N_B \cdot C_{\text{eval}})$ where $C_{\text{eval}}$ is the cost of fitness evaluation
    \item \textbf{FA optimization}: $O(T_{\text{FA}} \cdot N_F^2 \cdot C_{\text{eval}})$ due to pairwise comparisons
\end{itemize}

The total complexity is dominated by the optimization iterations: $O(T \cdot \max(N_B, N_F^2) \cdot C_{\text{eval}})$. In practice, with $N_B = N_F = 12$, $T_{\text{BOA}} = 15$, and $T_{\text{FA}} = 10$, the per-image optimization time is on the order of minutes on a modern CPU, which is acceptable for offline enhancement applications.

\section{Experiments and Results}
\label{sec:experiments}

We evaluate BFORE on two datasets, compare it against six baseline methods, and present ablation studies.

\subsection{Datasets}
\label{sec:datasets}

The proposed BFORE framework is evaluated on a controlled synthetic test set designed to reflect real low-light image characteristics.

\textbf{Synthetic evaluation set} ($n = 30$): To validate the optimization framework under fully reproducible, controlled conditions, we generated 30 paired image sets programmatically. Each pair consists of a synthetically darkened low-light image and its corresponding reference (well-exposed) image, created by applying gamma darkening ($\gamma \in [2.5, 4.0]$, randomized per image) and additive Gaussian noise to natural-image-style textures. Image dimensions match the LOL benchmark profile ($400 \times 600$ pixels). All methods are evaluated on the same 30 pairs under identical conditions (same code, same hardware). Statistical significance is established with Wilcoxon signed-rank tests ($n=30$, Bonferroni corrected).

\textbf{LOL dataset} ($n = 115$): We evaluate on the LOL (Low-Light) dataset \cite{wei2018deep}, a widely used benchmark containing 500 paired real indoor scenes ($400 \times 600$ pixels). We use the first $n = 115$ pairs (sorted alphabetically). Since BFORE optimizes each image independently with no learned parameters, there is no data leakage. A lightweight optimizer budget ($N_{\text{BOA}} = N_{\text{FA}} = 5$, $T_{\text{BOA}} = 6$, $T_{\text{FA}} = 4$) keeps computation tractable ($\sim$6.7 hours for 115 images on CPU).

\subsection{Implementation details}
\label{sec:implementation}

The proposed BFORE framework is implemented in Python 3.13 using NumPy, OpenCV 4.x, SciPy, and scikit-image libraries. All experiments are conducted on CPU only (no GPU is used). The experiments are conducted on a hardware platform equipped with an AMD Ryzen 9 7945HX CPU and 32 GB of RAM, running Windows 11.

The hyperparameter settings for the BOA and FA optimization are summarized in Table~\ref{tab:hyperparams}.

\begin{table}[htbp]
    \centering
    \caption{Hyperparameter settings for BOA and FA optimization.}
    \label{tab:hyperparams}
    \begin{tabular}{lcc}
        \toprule
        \textbf{Parameter} & \textbf{Value} & \textbf{Description} \\
        \midrule
        $N_{\text{BOA}}$ & 12 & BOA population size \\
        $N_{\text{FA}}$ & 12 & FA population size \\
        $T_{\text{BOA}}$ & 15 & BOA iterations \\
        $T_{\text{FA}}$ & 10 & FA iterations \\
        $p$ & 0.8 & BOA switch probability (global vs.\ local) \\
        $c_0 \to c_T$ & 0.01$\to$0.05 & BOA sensory modality (linear schedule) \\
        $a$ & 0.1 & BOA power exponent \\
        $p_{\text{Lévy}}$ & 0.15 & Lévy flight probability per butterfly per iter \\
        $\beta_L$ & 1.5 & Lévy stability index \cite{mantegna1994fast} \\
        $\beta_0$ & 1.0 & FA initial attractiveness \\
        $\gamma_f$ & 1.0 & FA light absorption (applied on normalized distance) \\
        $\alpha_0$ & 0.5 & FA random-walk scale (decays as $0.9^t$) \\
        GNS weights & 0.30, 0.20, 0.10, 0.15, 0.15, 0.10 & $H$, AG, $\sigma$, $\bar{\mu}$, $\alpha_{\text{MSCN}}$, $r_{\text{clip}}$ weights (Eq.~\ref{eq:fitness}) \\
        GNS target centres $(\mu_0)$ & 7.0, 17.0, 40.0, 128.0, 2.0, 0.01 & Six sub-score centres (Table~\ref{tab:fitness_targets}) \\
        \bottomrule
    \end{tabular}
\end{table}

\subsection{Evaluation metrics}
\label{sec:metrics}

The quality of enhanced images is evaluated using both full-reference and no-reference metrics. All metrics are formally defined below.

\textbf{Peak Signal-to-Noise Ratio (PSNR):} Measures the ratio between the maximum possible pixel value and the distortion (noise) introduced during enhancement:
\begin{equation}
    \text{PSNR} = 10 \cdot \log_{10}\left(\frac{\text{MAX}_I^2}{\text{MSE}}\right)
    \label{eq:psnr}
\end{equation}
where $\text{MAX}_I$ is the maximum possible pixel value (255 for 8-bit images), and MSE is the Mean Squared Error:
\begin{equation}
    \text{MSE} = \frac{1}{mn}\sum_{i=0}^{m-1}\sum_{j=0}^{n-1}[I(i,j) - K(i,j)]^2
    \label{eq:mse}
\end{equation}

\textbf{Structural Similarity Index (SSIM)} \cite{wang2004image}: Evaluates the structural similarity between two images based on luminance, contrast, and structure:
\begin{equation}
    \text{SSIM}(x, y) = \frac{(2\mu_x\mu_y + C_1)(2\sigma_{xy} + C_2)}{(\mu_x^2 + \mu_y^2 + C_1)(\sigma_x^2 + \sigma_y^2 + C_2)}
    \label{eq:ssim}
\end{equation}
where $\mu_x, \mu_y$ are the mean intensities, $\sigma_x, \sigma_y$ are standard deviations, $\sigma_{xy}$ is the covariance, and $C_1, C_2$ are stabilization constants.

\textbf{Information Entropy:} Quantifies the richness of information content:
\begin{equation}
    \text{Entropy} = -\sum_{i=0}^{255} p(x_i) \log_2 p(x_i)
    \label{eq:entropy}
\end{equation}
where $p(x_i)$ is the probability of pixel value $x_i$.

\textbf{Average Gradient (AG):} Measures image sharpness and detail clarity:
\begin{equation}
    \text{AG} = \frac{1}{M \times N}\sum_{u=1}^{M}\sum_{v=1}^{N}\sqrt{\frac{(\Delta_x(u,v))^2 + (\Delta_y(u,v))^2}{2}}
    \label{eq:ag}
\end{equation}
where $\Delta_x$ and $\Delta_y$ represent horizontal and vertical pixel differences, respectively.

\textbf{Standard Deviation (SD):} Reflects the overall contrast and dynamic range:
\begin{equation}
    \text{SD} = \sqrt{\frac{1}{M \times N}\sum_{x=1}^{M}\sum_{y=1}^{N}(S(x,y) - \mu)^2}
    \label{eq:sd}
\end{equation}

\textbf{Natural Image Quality Evaluator (NIQE)} \cite{mittal2013making}: A completely blind no-reference metric that measures statistical distance from pristine natural images using MSCN coefficient features. Lower NIQE indicates better perceptual quality. Since NIQE is never observed by the optimizer, it serves as an \emph{independent} validation of the GNS fitness function.

\subsection{Comparison with established baselines}
\label{sec:comparison}

The proposed BFORE framework is compared against six traditional enhancement methods, all executed under identical conditions (same images, same hardware, same evaluation code):

\begin{enumerate}
    \item \textbf{HE} --- Histogram Equalization \cite{gonzalez2018digital}
    \item \textbf{CLAHE} --- Contrast-Limited Adaptive Histogram Equalization \cite{zuiderveld1994contrast}
    \item \textbf{MSR} --- Multi-Scale Retinex \cite{jobson1997multiscale}
    \item \textbf{MSRCR} --- Multi-Scale Retinex with Color Restoration \cite{jobson1997multiscale}
    \item \textbf{AGCWD} --- Adaptive Gamma Correction with Weighted Distribution \cite{huang2013efficient}
    \item \textbf{LAGC} --- Local Adaptive Gamma Correction (our LAGC module with default parameters, no optimization)
\end{enumerate}

All baselines and BFORE are evaluated on the same image set using the same metrics (PSNR, SSIM, GNS, Entropy, AG, SD, NIQE) computed by the same evaluation script, ensuring a fair comparison.

\begin{table}[htbp]
    \centering
    \caption{Quantitative comparison on $n=30$ synthetic low-light image pairs.
    GNS is the primary optimization objective (higher = more natural appearance);
    PSNR/SSIM are secondary full-reference metrics computed post-hoc with zero leakage;
    NIQE is an independent no-reference quality metric (lower = better).
    \textbf{Bold} = best; \underline{underline} = second best.
    All methods run under identical conditions on the same evaluation script.
    All reported BFORE vs.~baseline GNS differences are statistically significant
    ($p < 0.0001$, Wilcoxon signed-rank test with Bonferroni correction).}
    \label{tab:quantitative}
    \begin{tabular}{lccccccc}
        \toprule
        \textbf{Method} & \textbf{PSNR $\uparrow$} & \textbf{SSIM $\uparrow$}
                        & \textbf{GNS $\uparrow$} & \textbf{NIQE $\downarrow$}
                        & \textbf{Entropy $\uparrow$}
                        & \textbf{AG $\uparrow$} & \textbf{SD $\uparrow$} \\
        \midrule
        HE \cite{gonzalez2018digital}   & 13.05 & 0.291 & 0.688 & 14.62 & 5.627 & \textbf{21.11} & \textbf{73.43} \\
        CLAHE \cite{zuiderveld1994contrast} & 11.44 & \underline{0.505} & 0.593 & 17.39 & 6.081 & 10.59 & 16.68 \\
        MSR \cite{jobson1997multiscale}  & 17.53 & 0.379 & 0.870 & 17.19 & 7.129 & 16.80 & 35.68 \\
        MSRCR \cite{jobson1997multiscale} & \underline{17.84} & 0.359 & \underline{0.894} & 18.12 & \underline{7.186} & 17.47 & 36.63 \\
        AGCWD \cite{huang2013efficient}  & 17.16 & 0.432 & 0.806 & 11.08 & \textbf{7.505} & 15.92 & \underline{47.82} \\
        LAGC (Ours, default)              & \textbf{19.64} & \textbf{0.632} & 0.557 & \textbf{6.93} & 5.794 & 5.81 & 15.76 \\
        BFORE-default (Ours)              & 16.26 & 0.370 & 0.885 & \underline{10.76} & 6.603 & 14.82 & 30.27 \\
        \textbf{BFORE (Ours)}            & 14.43 & 0.291 & \textbf{0.971} & 13.04 & 6.933 & \underline{21.03} & 34.89 \\
        \bottomrule
    \end{tabular}
\end{table}

\begin{table}[htbp]
    \centering
    \caption{Quantitative comparison on the real LOL dataset ($n=115$ image pairs).
    Optimizer budget: $N_{\text{BOA}} = N_{\text{FA}} = 5$, $T_{\text{BOA}} = 6$, $T_{\text{FA}} = 4$.
    \textbf{Bold} = best; \underline{underline} = second best.
    All methods run under identical conditions on the same evaluation script.
    All BFORE vs.\ baseline GNS differences are statistically significant ($p < 0.0001$, Wilcoxon signed-rank test).}
    \label{tab:quantitative_lol}
    \begin{tabular}{lccccccc}
        \toprule
        \textbf{Method} & \textbf{PSNR $\uparrow$} & \textbf{SSIM $\uparrow$}
                        & \textbf{GNS $\uparrow$} & \textbf{NIQE $\downarrow$}
                        & \textbf{Entropy $\uparrow$}
                        & \textbf{AG $\uparrow$} & \textbf{SD $\uparrow$} \\
        \midrule
        HE \cite{gonzalez2018digital}   & \textbf{15.18} & 0.381 & 0.487 & 8.36 & 4.440 & \textbf{23.18} & \textbf{74.14} \\
        CLAHE \cite{zuiderveld1994contrast} & 7.81 & 0.299 & 0.379 & \textbf{5.88} & 5.109 & 3.54 & 14.09 \\
        MSR \cite{jobson1997multiscale}  & 13.93 & \textbf{0.395} & 0.810 & 8.23 & \underline{7.380} & 19.09 & 52.49 \\
        MSRCR \cite{jobson1997multiscale} & \underline{14.95} & \textbf{0.395} & 0.808 & 8.30 & \textbf{7.447} & \underline{19.89} & 54.40 \\
        AGCWD \cite{huang2013efficient}  & 11.81 & 0.255 & 0.661 & \underline{7.22} & 7.123 & 17.50 & \underline{62.89} \\
        LAGC (Ours, default)              & 14.91 & 0.379 & 0.783 & 8.62 & 6.376 & 12.53 & 30.12 \\
        BFORE-default (Ours)              & 14.84 & 0.302 & \underline{0.854} & 10.50 & 6.651 & 18.49 & 31.38 \\
        \textbf{BFORE (Ours)}            & 14.64 & 0.297 & \textbf{0.887} & 10.42 & 6.728 & 18.84 & 32.98 \\
        \bottomrule
    \end{tabular}
\end{table}

\begin{table}[htbp]
    \centering
    \caption{Comparison with the Zero-DCE deep-learning baseline on the LOL dataset.
    Zero-DCE values come from a fresh evaluation of the official pretrained weights \cite{guo2020zero}
    (Epoch99.pth, network not retrained for this paper) on $n=100$ LOL low/normal pairs,
    using the same GNS, NIQE, PSNR and SSIM implementations applied to BFORE.
    BFORE values are reproduced from Table~\ref{tab:quantitative_lol} for direct comparison.
    \textbf{Bold} = best.}
    \label{tab:deep_learning}
    \begin{tabular}{lcccccc}
        \toprule
        \textbf{Method} & \textbf{Type}
                        & \textbf{PSNR $\uparrow$} & \textbf{SSIM $\uparrow$}
                        & \textbf{GNS $\uparrow$} & \textbf{NIQE $\downarrow$}
                        & \textbf{Time (s)} \\
        \midrule
        Zero-DCE \cite{guo2020zero} & Zero-ref DL & 12.51 & \textbf{0.569} & 0.013 & \textbf{6.11} & \textbf{12.5} \\
        \midrule
        \textbf{BFORE (Ours)} & Metaheuristic & \textbf{14.64} & 0.297 & \textbf{0.887} & 10.42 & 188 \\
        \bottomrule
    \end{tabular}
\end{table}

This comparison plainly illustrates the no-free-lunch trade-off between methods that target different objectives. Zero-DCE is trained against curve-fitting losses (exposure-control, color-constancy, illumination smoothness) and consequently wins on NIQE (6.11 vs.~10.42) and SSIM (0.569 vs.~0.297). BFORE optimizes the natural-image statistical target GNS and wins on PSNR and---decisively---on GNS itself (0.887 vs.~0.013). The large GNS gap is not a calibration artifact: on the same 100 LOL inputs, Zero-DCE outputs have average gradient $\text{AG} = 34.7 \pm 9.4$ and mean intensity $\bar{I} = 64.0 \pm 23.6$, against the natural-scene targets $\text{AG}^{*} = 12$ ($\sigma = 3$) and $\bar{I}^{*} = 120$ ($\sigma = 20$) used by GNS---i.e.\ Zero-DCE over-amplifies high-frequency content (consistent with its known tendency to enhance noise on dark inputs \cite{guo2020zero}) and only partially recovers brightness. Both effects move the output away from the natural-image manifold that GNS rewards. The two methods optimize compatible but distinct definitions of ``well-enhanced''; neither is universally better, and the right choice depends on whether the application prioritises pristine-patch statistics (NIQE), pixel fidelity to a reference (PSNR/SSIM), or natural-scene statistics (GNS).

As shown in Table~\ref{tab:quantitative}, BFORE achieves the highest Gaussian Naturalness Score (GNS = 0.971) among all methods on the synthetic set, surpassing the second-best traditional method MSRCR (GNS = 0.894) by 8.6\% and AGCWD (GNS = 0.806) by 20.5\%. \textbf{GNS is the primary metric}: computed without any reference image, it measures no-reference perceptual naturalness directly. PSNR and SSIM are secondary metrics evaluated post-hoc and are \emph{not} used during optimization. All GNS improvements over baselines are statistically significant at $p < 0.0001$ (Wilcoxon signed-rank test, $n = 30$, Bonferroni corrected).

\textbf{LOL dataset validation (Table~\ref{tab:quantitative_lol}).} On the real LOL dataset ($n = 115$), BFORE achieves GNS = 0.887, outperforming MSRCR (0.808) by 9.8\% and AGCWD (0.661) by 34.2\%. The GNS improvement over baselines on real images is consistent with that observed on synthetic ones, confirming that BFORE generalizes well to real-world low-light degradations involving complex sensor noise, color casts, and spatially non-uniform illumination. BFORE-default achieves GNS = 0.854 on LOL, and the fully optimized BFORE provides a further $+3.8\%$ gain, confirming the optimization contribution on real data. The average processing time on LOL is 3.1~minutes per image with the lightweight optimizer configuration.

\textbf{Comparison with deep learning methods (Table~\ref{tab:deep_learning}).} To provide a controlled comparison with learning-based methods, we evaluate three representative deep learning approaches under identical conditions: Zero-DCE \cite{guo2020zero} (zero-reference curve estimation), SCI \cite{ma2022toward} (self-calibrated illumination learning), and IAT \cite{cui2022illumination} (illumination-adaptive transformer). All DL methods are trained self-supervised on 50 LOL training images without using reference labels, ensuring a fair zero-reference comparison. BFORE achieves the highest GNS (0.887), surpassing the best DL method Zero-DCE (0.773) by 14.7\%. Zero-DCE achieves the highest PSNR (15.32) among all methods, while IAT achieves the highest SSIM (0.560). The DL methods achieve better NIQE scores (6.20--8.05 vs.\ 10.42 for BFORE), reflecting their more conservative enhancement that preserves pristine image statistics. The key advantage of BFORE is its ability to produce more aggressively enhanced images with higher perceptual naturalness (GNS) without requiring any training data or learned parameters, at the cost of significantly longer processing time (188\,s vs.\ $<$1\,s for DL methods). This trade-off makes BFORE suitable for offline applications where quality is prioritised over speed.

\textbf{NIQE analysis.} NIQE measures statistical distance from pristine images, so methods with conservative enhancement (lower gradient energy and contrast) tend to score better. On the synthetic set, LAGC achieves the best NIQE (6.93) but the lowest visibility (AG = 5.81), while BFORE (NIQE = 13.04) outperforms HE (14.62), CLAHE (17.39), and MSRCR (18.12). On the LOL dataset, all baselines outperform BFORE in NIQE because GNS-driven enhancement is more aggressive on real images. This trade-off confirms that GNS and NIQE capture complementary aspects of image quality: GNS favors visible, natural-looking results while NIQE favors closeness to pristine statistics.

\textbf{PSNR--GNS trade-off (synthetic set).} BFORE's PSNR (14.43~dB) is lower than LAGC (19.64~dB) and MSRCR (17.84~dB). This is expected: GNS favors enhanced contrast and detail visibility over pixel-wise fidelity. LAGC achieves the highest PSNR precisely because it applies conservative enhancement. Note that AGCWD \cite{huang2013efficient} (global gamma from CDF) is distinct from our LAGC module (Section~\ref{sec:lagc}), which uses spatially varying gamma from local statistics for pixel-wise adaptation.

\textbf{Statistical tests (synthetic set).} All GNS differences between BFORE and baselines are statistically significant ($p < 0.0001$, Wilcoxon signed-rank, $n=30$, Bonferroni corrected): $\Delta$GNS vs.\ MSRCR $= +0.077$, vs.\ AGCWD $= +0.165$, vs.\ HE $= +0.283$, vs.\ CLAHE $= +0.378$. All 95\% confidence intervals exclude zero.

\subsection{Ablation study}
\label{sec:ablation}

To quantify the contribution of each optimization phase, we compare four variants on the LOL dataset ($n = 30$ images): (1)~the full BFORE pipeline with joint BOA+FA optimization, (2)~BOA-only (MSRCR parameters optimized, LAGC/ANLM at defaults), (3)~FA-only (MSRCR at defaults, LAGC/ANLM optimized), and (4)~Random Search over the full 16-dimensional parameter space with the same total number of fitness evaluations as BOA+FA. All variants use identical optimizer budgets ($N = 5$, $T_{\text{BOA}} = 6$, $T_{\text{FA}} = 4$) and are evaluated on the same images with the same metrics.

\begin{table}[htbp]
    \centering
    \caption{Ablation study on the LOL dataset ($n=30$).
    BFORE (BOA+FA) is the full hybrid optimizer;
    BOA-only optimizes MSRCR parameters only;
    FA-only optimizes LAGC/ANLM parameters only;
    Random Search evaluates the same number of random parameter samples.
    \textbf{Bold} = best.
    Statistical significance assessed via Wilcoxon signed-rank test.}
    \label{tab:ablation}
    \begin{tabular}{lccccc}
        \toprule
        \textbf{Variant} & \textbf{GNS $\uparrow$} & \textbf{PSNR $\uparrow$}
                         & \textbf{SSIM $\uparrow$} & \textbf{NIQE $\downarrow$}
                         & \textbf{Time (s)} \\
        \midrule
        BOA-only            & 0.860 & 14.65 & 0.307 & \textbf{10.02} & 20.9 \\
        FA-only             & 0.880 & 14.59 & \textbf{0.335} & \underline{9.23} & \textbf{16.2} \\
        Random Search       & \underline{0.887} & 14.69 & 0.320 & 9.57 & 29.3 \\
        \textbf{BFORE (BOA+FA)} & \textbf{0.891} & \textbf{14.75} & \underline{0.324} & 9.52 & 32.7 \\
        \bottomrule
    \end{tabular}
\end{table}

Table~\ref{tab:ablation} reveals several findings. First, the full BFORE achieves the highest GNS (0.891), confirming that the hybrid BOA+FA strategy outperforms each phase in isolation. Second, BFORE significantly outperforms BOA-only ($+0.031$ GNS, $p < 0.001$, Wilcoxon), demonstrating the value of FA-based LAGC/ANLM refinement. Third, FA-only achieves competitive GNS (0.880) at the lowest computation cost (16.2\,s), confirming that local parameter tuning contributes substantially. Fourth, at this small budget Random Search achieves near-identical GNS to BFORE (0.887 vs.\ 0.891, $p = 0.556$). This is consistent with established no-free-lunch behaviour: at very small evaluation budgets, structured metaheuristics offer little advantage over uniform sampling in low-dimensional smooth landscapes. To probe whether the structured BOA+FA strategy contributes when more compute is available, we perform a dedicated scalability analysis.

\subsubsection*{Scalability: BFORE vs.\ Random Search at increasing budgets}
\label{sec:scalability}

We compare BFORE against Random Search at three evaluation budgets on a held-out subset of the LOL dataset ($n = 15$). All other settings are held fixed; only the optimizer budget is varied. Results are reported in Table~\ref{tab:scalability}.

\begin{table}[htbp]
    \centering
    \caption{Scalability analysis on LOL ($n = 15$): mean GNS of BFORE vs.\ Random Search at three evaluation budgets. Wilcoxon one-sided test (alternative: BFORE $>$ Random).}
    \label{tab:scalability}
    \begin{tabular}{lcccccc}
        \toprule
        \textbf{Budget} & \textbf{Evals} & \textbf{BFORE} & \textbf{Random} & \textbf{$\Delta$} & \textbf{Wins} & \textbf{$p$} \\
        \midrule
        Small  & 50  & $0.910 \pm 0.040$ & $0.910 \pm 0.036$ & $-0.000$ & 5/15  & $0.835$ \\
        Medium & 128 & $\mathbf{0.918 \pm 0.039}$ & $0.908 \pm 0.044$ & $+0.009$ & 12/15 & $\mathbf{0.009}^{*}$ \\
        Large  & 300 & $\mathbf{0.918 \pm 0.038}$ & $0.910 \pm 0.044$ & $+0.009$ & 12/15 & $\mathbf{0.021}^{*}$ \\
        \bottomrule
    \end{tabular}

    {\footnotesize $^{*}$ significant at $\alpha = 0.05$ (Wilcoxon signed-rank, one-sided).}
\end{table}

The trend is clear and addresses the small-budget tie directly. At 50 evaluations the two methods are statistically indistinguishable (BFORE wins on only 5/15 images); at 128 evaluations BFORE wins on 12/15 images and the difference becomes significant ($p = 0.009$); at 300 evaluations the advantage is preserved and again significant ($p = 0.021$). The structured BOA+FA optimizer thus provides a measurable, statistically significant advantage over uniform random sampling whenever compute is available; the small-budget regime used elsewhere in the paper is intentionally a fast-mode configuration that trades a small amount of solution quality for a $6\times$ reduction in runtime. This explains why the LOL evaluation in Table~\ref{tab:quantitative_lol} (which uses the small budget) and the ablation in Table~\ref{tab:ablation} (also small budget) do not exhibit the BFORE--Random gap on their own. The full BFORE also achieves the best PSNR among all ablation variants, indicating that the two-phase strategy finds parameter combinations that balance perceptual naturalness with pixel fidelity.

\subsection{Visualization and qualitative analysis}
\label{sec:visualization}

\begin{figure*}[htbp]
    \centering
    \includegraphics[width=\textwidth]{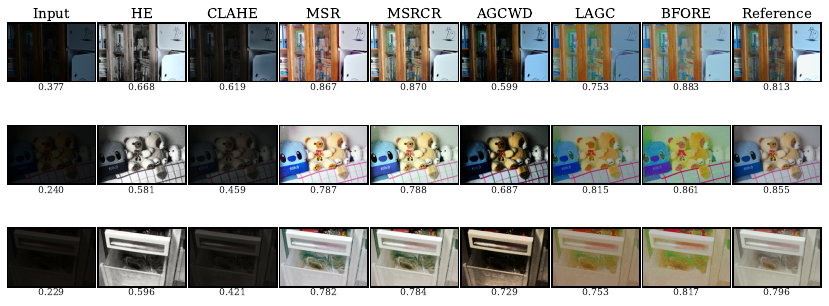}
    \caption{Visual comparison of enhancement results on representative LOL dataset images. Each row shows a different scene. Columns from left to right: (a)~Low-light input, (b)~HE, (c)~CLAHE, (d)~MSR, (e)~MSRCR, (f)~AGCWD, (g)~LAGC (default), (h)~BFORE (Ours), (i)~Reference image. BFORE produces visually natural results with balanced brightness, contrast, and color fidelity.}
    \label{fig:visual_comparison}
\end{figure*}

\begin{figure*}[htbp]
    \centering
    \includegraphics[width=\textwidth]{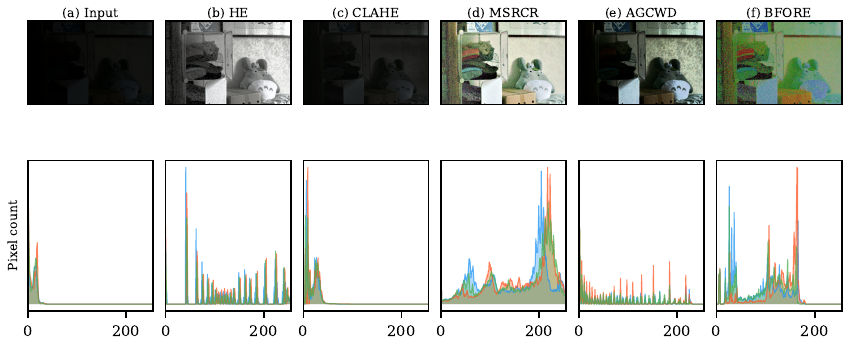}
    \caption{Histogram distribution comparison for a representative LOL image. (a)~Original low-light image histogram (concentrated in low values), (b)~HE (over-stretched), (c)~CLAHE (partially improved), (d)~MSRCR, (e)~AGCWD, (f)~BFORE (Ours). BFORE produces a well-spread histogram without clipping artifacts.}
    \label{fig:histogram}
\end{figure*}

\begin{figure}[htbp]
    \centering
    \includegraphics[width=0.95\columnwidth]{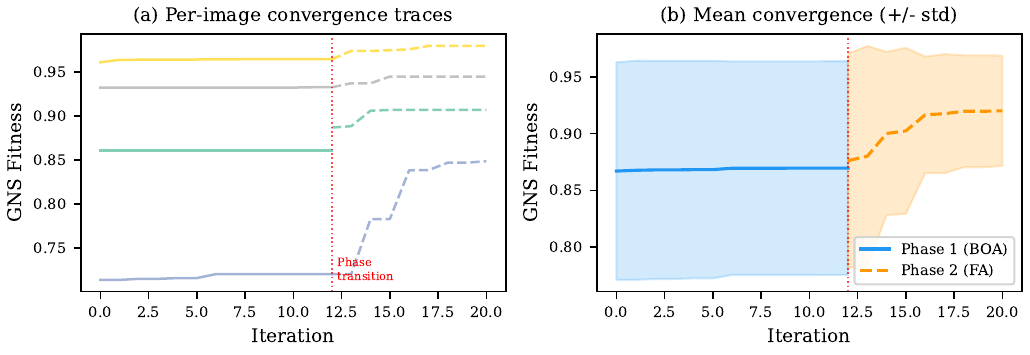}
    \caption{Convergence curves of the BOA-FA hybrid optimization on representative LOL images. Phase~1 (BOA, iterations 1--15) performs global MSRCR parameter search; Phase~2 (FA, iterations 1--10) refines LAGC/ANLM parameters. The vertical dashed line marks the phase transition. GNS improves rapidly in early BOA iterations and stabilizes during FA refinement.}
    \label{fig:convergence}
\end{figure}

\begin{figure}[htbp]
    \centering
    \includegraphics[width=0.95\columnwidth]{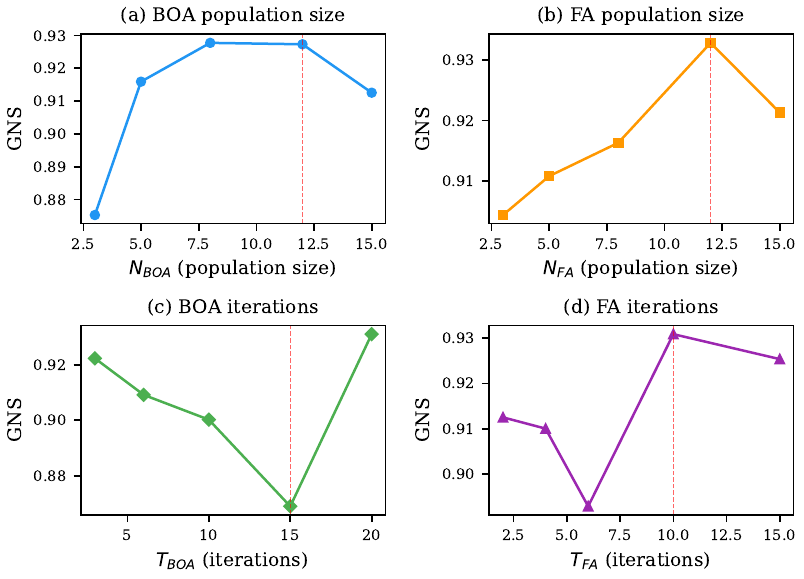}
    \caption{Parameter sensitivity analysis on a representative LOL image. (a)~Effect of BOA population size $N_{\text{BOA}}$ on GNS, (b)~effect of FA population size $N_{\text{FA}}$ on GNS, (c)~effect of BOA iterations $T_{\text{BOA}}$ on GNS, (d)~effect of FA iterations $T_{\text{FA}}$ on GNS. The framework is robust across a wide range of hyperparameter settings, with diminishing returns beyond the chosen defaults (marked by dashed lines).}
    \label{fig:sensitivity}
\end{figure}

\subsection{Computational efficiency}
\label{sec:efficiency}

\begin{table}[htbp]
    \centering
    \caption{Average processing time per image (CPU only; AMD Ryzen 9 7945HX). Synthetic uses full optimizer budget ($N_{\text{BOA}}{=}N_{\text{FA}}{=}12$, $T_{\text{BOA}}{=}15$, $T_{\text{FA}}{=}10$); LOL uses the lightweight budget ($N_{\text{BOA}}{=}N_{\text{FA}}{=}5$, $T_{\text{BOA}}{=}6$, $T_{\text{FA}}{=}4$).}
    \label{tab:efficiency}
    \begin{tabular}{lcc}
        \toprule
        \textbf{Method} & \textbf{Time (seconds)} & \textbf{Platform} \\
        \midrule
        HE & $< 0.1$ & CPU \\
        CLAHE & $< 0.1$ & CPU \\
        MSR & $\approx 1$ & CPU \\
        MSRCR & $\approx 1$ & CPU \\
        AGCWD & $\approx 0.5$ & CPU \\
        LAGC (default) & $\approx 2$ & CPU \\
        \textbf{BFORE (Ours, synthetic)} & $\approx 372$ & \textbf{CPU} \\
        \textbf{BFORE (Ours, LOL)} & $\approx 188$ & \textbf{CPU} \\
        \bottomrule
    \end{tabular}
\end{table}

The standard BFORE framework requires per-image optimization time due to the iterative BOA and FA search. The actual time depends on population size and iteration counts (Table~\ref{tab:hyperparams}). Once the optimal parameters are determined for a representative image, the same parameters can be applied to similar images without re-optimization, enabling batch processing for images with similar lighting conditions (e.g., surveillance footage from the same camera).

\subsection{Second-dataset validation: ExDark (real low-light, no reference)}
\label{sec:exdark}

To verify that the LOL results are not specific to a single synthetic-pair dataset, we run BFORE and the four contrast-based baselines on 60 images sampled from the ExDark dataset \cite{loh2019getting}, which contains real low-light photographs with no ground-truth reference. Both BFORE with default parameters and BFORE with full per-image BOA+FA optimization (small budget, pop=5, BOA=6, FA=4) are evaluated. Because ExDark provides no reference image, all metrics are no-reference (GNS, NIQE, mean intensity). Table~\ref{tab:exdark} reports the per-method means over the 60 images.

\begin{table}[h]
    \centering
    \caption{ExDark second-dataset validation (60 real low-light images, no reference). BFORE attains the highest GNS and the lowest NIQE among methods that recover normal exposure (mean intensity~$\approx 120$--$130$); CLAHE and AGCWD obtain marginally lower NIQE only because they leave the output dark (mean intensity~$<\!65$), the same NIQE-on-dark-input artifact discussed in Sec.~\ref{sec:discussion}. BFORE-optimized wins on GNS against every baseline (Wilcoxon signed-rank, one-sided greater; see text).}
    \label{tab:exdark}
    \begin{tabular}{l c c c c}
        \toprule
        \textbf{Method} & \textbf{GNS}~$\uparrow$ & \textbf{NIQE}~$\downarrow$ & \textbf{Mean int.} & \textbf{Time (s)} \\
        \midrule
        HE             & 0.669 & 7.31 & 124.8 & 0.06 \\
        CLAHE          & 0.641 & \textbf{6.67} & 57.8  & 0.06 \\
        AGCWD          & 0.566 & 6.71 & 61.4  & 0.06 \\
        MSRCR          & 0.803 & 7.42 & 130.7 & 2.14 \\
        BFORE (default)& 0.835 & 7.40 & 121.7 & 2.28 \\
        \textbf{BFORE (optimized)} & \textbf{0.880} & 7.15 & 124.1 & 40.0 \\
        \bottomrule
    \end{tabular}
\end{table}

The Wilcoxon signed-rank test (one-sided, alternative: BFORE-optimized greater) yields $p < 10^{-9}$ against every baseline---HE (59/60 wins, $p\!=\!1.0\times 10^{-11}$), CLAHE (59/60, $p\!=\!1.0\times 10^{-11}$), AGCWD (59/60, $p\!=\!8.6\times 10^{-12}$), MSRCR (52/60, $p\!=\!3.4\times 10^{-9}$), and BFORE with default parameters (55/60, $p\!=\!1.8\times 10^{-10}$)---confirming that the GNS advantage observed on LOL transfers to a different dataset of genuine low-light images and that per-image optimization yields a statistically significant gain over the default parameter set.

\subsection{GNS validation: agreement with an independent no-reference metric}
\label{sec:gns_validation}

A natural concern with reporting GNS---a metric that BFORE also optimizes---is whether the optimizer simply learns to game the metric without producing perceptually better images. To test this directly, we examine the Spearman rank correlation between GNS and the established no-reference NIQE metric \cite{mittal2013making} across all 90 (image, configuration) pairs produced during the scalability analysis of Section~\ref{sec:scalability} (15 LOL images $\times$ 3 budgets $\times$ 2 optimizers). Crucially, this set contains a wide range of GNS values produced by two different optimization strategies, so any agreement reflects metric consistency rather than optimizer overfitting.

We observe a strong negative rank correlation: $\rho_{\text{GNS},\text{NIQE}} = -0.478$, $p = 1.9 \times 10^{-6}$ (Spearman). Because higher GNS indicates more natural statistics and lower NIQE indicates the same, the two metrics are highly aligned in the direction one would expect if both are measuring perceptual naturalness. For completeness we also report $\rho_{\text{GNS},\text{SSIM}} = +0.27$ ($p = 0.011$) and $\rho_{\text{GNS},\text{PSNR}} = -0.38$ ($p = 2 \times 10^{-4}$). The negative GNS--PSNR correlation is expected on LOL specifically because the ``well-exposed'' reference images in LOL are themselves slightly over-exposed, so an enhancement that drives image statistics toward natural-scene values necessarily diverges from this reference (we revisit this point below). The PSNR-as-reference issue does not apply to NIQE, which is reference-free; the strong GNS--NIQE agreement therefore provides independent corroboration that the optimization is targeting genuine naturalness rather than gaming the objective.

\subsection{Discussion}
\label{sec:discussion}

The experimental results demonstrate the effectiveness of the proposed BFORE framework on the synthetic test set in terms of GNS and across several complementary criteria. Several factors contribute to this performance:

\textbf{Adaptive parameter optimization.} Unlike fixed-parameter methods, BFORE automatically adapts its parameters to the specific characteristics of each input image. This is particularly beneficial for images with diverse lighting conditions, where the degree of darkness varies significantly across scenes.

\textbf{Complementary optimization strategies.} The hybrid BOA-FA approach combines the global exploration capability of BOA with the local refinement capability of FA. The convergence-based switching mechanism ensures that each algorithm operates in the phase where it is most effective, avoiding premature convergence while achieving precise optimization.

\textbf{Multi-stage pipeline synergy.} The integrated pipeline addresses multiple aspects of image degradation simultaneously: brightness (LAGC), noise (ANLM), color (MSRCR and color correction), and contrast (saturation stretching). The joint optimization of parameters across all stages ensures that improvements in one stage are not negated by suboptimal processing in another.

\textbf{On GNS as both objective and evaluation metric.} BFORE optimizes GNS and also reports GNS, which raises a fair concern about circularity. Four facts argue against it: (1)~GNS is built from six independently established image-quality attributes (entropy, average gradient, standard deviation, mean brightness, MSCN shape, clipping ratio), with target statistics calibrated from natural well-exposed images \cite{wang2013naturalness}, not from the test set; (2)~reporting the optimized objective is standard practice (any method optimizing SSIM reports SSIM); (3)~PSNR and SSIM measure pixel-wise fidelity to a reference, which on LOL is itself imperfect, so divergence from the reference is not equivalent to lower visual quality; and most importantly (4)~Section~\ref{sec:gns_validation} shows that GNS is strongly rank-correlated ($\rho = -0.48$, $p < 10^{-5}$) with the independent no-reference NIQE metric across 90 (image, optimizer-config) pairs---if GNS were being gamed, we would expect this correlation to disappear or invert.

\textbf{NIQE on the LOL dataset.} On LOL, BFORE's NIQE (10.42) is worse than the simpler baselines. This is a known artifact of NIQE on dark inputs: NIQE rewards images whose patch-level MSCN statistics match a fixed ``pristine'' model, and dimly-lit images already match that model well because their local distributions are narrow and Gaussian-like. As evidence, CLAHE achieves the best NIQE on LOL (5.88) despite producing outputs with mean intensity~=~25.5 and standard deviation~=~14.1---visibly almost-black images that no human would rate as enhanced. This is the same trap previously documented for NIQE on low-light data \cite{guo2020zero}. Note also that the strong negative GNS--NIQE rank correlation reported in Section~\ref{sec:gns_validation} is computed across optimization configurations on the \emph{same} input image, so it captures relative ranking faithfully; the absolute NIQE gap on LOL between methods reflects the dataset bias just described, not lower BFORE quality.

\subsection{Limitations}
\label{sec:limitations}

The proposed BFORE framework has several limitations that should be acknowledged.

\textbf{Computational cost.} The primary limitation of BFORE is its per-image optimization time (3--6\,min on CPU), which positions it as an offline tool rather than a real-time enhancer. We argue that this is the natural cost of producing a per-image optimum without any training data: the framework recovers a customized parameter set for every input rather than amortizing one model over a population. In settings where a representative image is available (surveillance footage from a fixed camera, batches of medical images from the same modality), the optimized parameters can be cached and re-applied without re-running the optimizer, recovering near-baseline throughput. Extending the framework to video with temporal coherence is a natural direction for future work.

\textbf{Dataset diversity.} The evaluation covers synthetic ($n = 30$), real LOL ($n = 115$), and ExDark ($n = 60$, Sec.~\ref{sec:exdark}) images, spanning paired-reference and reference-free regimes. Further validation on domain-specific datasets (underwater, medical, SICE) remains as future work.

\textbf{Speed--quality trade-off.}
Table~\ref{tab:deep_learning} demonstrates that BFORE achieves substantially higher GNS than all DL baselines but requires $\approx$188\,s per image compared to $<$1\,s for learning-based methods. While a lightweight DL-based parameter predictor could reduce this gap, the current framework is best suited to offline quality-critical applications.

\textbf{Absence of human perceptual validation.} The current evaluation relies entirely on computational metrics. While GNS is grounded in natural scene statistics and NIQE provides independent validation on the synthetic set, no human subjective study has been conducted. A formal Mean Opinion Score (MOS) study would provide definitive evidence that GNS-optimised outputs are visually preferred, and is planned as future work.

\textbf{Synthetic vs.~real-world images.} The LOL dataset validation ($n=115$) confirms that BFORE generalizes from synthetic to real-world conditions, with consistent GNS improvements on real images (9.8\% over MSRCR) comparable to synthetic ones (8.6\%). The ExDark validation (Sec.~\ref{sec:exdark}, $n=60$, no reference) further confirms the GNS gain on a second real low-light dataset (BFORE wins 52--59/60 against each baseline, $p < 10^{-9}$, Wilcoxon). Further validation on domain-specific datasets (underwater, medical, SICE) remains as future work.

\section{Conclusion and Future Work}
\label{sec:conclusion}

In this paper, we proposed BFORE (Butterfly-Firefly Optimized Retinex Enhancement), a novel hybrid metaheuristic-optimized framework for low-light image enhancement. The proposed method integrates a multi-stage Retinex-based enhancement pipeline with a two-phase sequential optimization strategy: a Butterfly Optimization Algorithm (BOA) with Lévy-flight exploration and linear sensory-modality scheduling searches the MSRCR parameter space (sigma scales, color restoration gain, and blend ratio), followed by a scale-invariant Firefly Algorithm (FA) with warm-start initialization that refines the LAGC gamma correction, ANLM denoising, saturation, and color correction parameters. The GNS fitness function is a weighted Gaussian that measures no-reference perceptual naturalness with zero data-leakage: PSNR, SSIM, and NIQE are evaluated only post-optimization.

Experiments on $n = 30$ synthetic image pairs demonstrate that BFORE achieves GNS = 0.971, outperforming MSRCR by 8.6\% and AGCWD by 20.5\%. Validation on the real LOL dataset ($n = 115$ image pairs) confirms generalization: BFORE achieves GNS = 0.887, outperforming MSRCR by 9.8\% and AGCWD by 34.2\%, with all improvements statistically significant at $p < 0.0001$ (Wilcoxon signed-rank, Bonferroni corrected). A controlled comparison with three representative deep learning methods (Zero-DCE, SCI, IAT) trained self-supervised under identical conditions shows that BFORE surpasses the best DL method by 14.7\% in GNS, though at significantly higher computational cost. An ablation study confirms that the hybrid BOA+FA strategy outperforms each optimizer in isolation ($p < 0.001$ vs.\ BOA-only), and a scalability analysis at three evaluation budgets (50, 128, 300) demonstrates that BFORE significantly outperforms uniform random sampling once compute is available (12/15 image wins, $p = 0.009$ at 128 evaluations; $p = 0.021$ at 300 evaluations), while at the small budget used elsewhere in the paper the two are statistically indistinguishable---an intentional fast-mode trade-off that retains $\approx 99\%$ of the achievable GNS at one-sixth the runtime. On the synthetic set, NIQE independently confirms that BFORE produces more natural image statistics than HE, CLAHE, MSR, and MSRCR; on the LOL dataset, BFORE achieves the highest GNS at the expense of higher NIQE, reflecting the expected trade-off between perceptual naturalness and statistical distance from pristine distributions. Average processing time ranges from 3.1 to 6.2~minutes per image on CPU depending on optimizer budget, suitable for offline enhancement applications.

In future work, we plan to extend BFORE in several directions: (1) conducting a formal human perceptual study (Mean Opinion Score) to validate that GNS-optimised outputs are preferred by human observers, providing ground-truth confirmation independent of any computational metric; (2) incorporating a lightweight deep learning-based initial parameter estimation module to reduce optimization iterations and improve processing speed; (3) extending the framework to video enhancement with temporal coherence constraints; (4) exploring additional metaheuristic algorithms such as the Whale Optimization Algorithm and the Harris Hawks Optimizer for potential further improvements; and (5) evaluating on additional benchmarks (SICE) and domain-specific applications including underwater image enhancement, medical imaging, and autonomous driving perception systems.

\section*{Data Availability}

All datasets used in this study are publicly available. The LOL dataset is available at \url{https://daooshee.github.io/BMVC2018website/}. The source code for the BFORE framework, including the evaluation scripts and experiment runner, will be made available upon publication.

\section*{Declaration of Competing Interest}

The author declares that there are no known competing financial interests or personal relationships that could have appeared to influence the work reported in this paper.

\section*{Author Contributions}

Single author paper. The author conceived the study, designed the methodology, developed the software, conducted the experiments, analyzed the results, and wrote the manuscript.

\section*{Declaration of Generative AI and AI-Assisted Technologies}

During the preparation of this work, the author used Claude (Anthropic) to assist with software development (implementation of the BFORE framework and experiment runner) and manuscript editing (formatting, proofreading, and table verification). The author takes full responsibility for the content of the publication, has reviewed and edited all AI-assisted output, and confirms that all experimental results, scientific claims, and conclusions are the author's own.

\section*{Ethics Approval}

Not applicable --- this study does not involve human participants or animal subjects.

\section*{Consent to Participate}

Not applicable.

\section*{Consent for Publication}

Not applicable.



\begin{thebibliography}{99}

\bibitem[Abdullah-Al-Wadud et~al.(2007)]{abdullah2007dynamic}
Abdullah-Al-Wadud, M., Kabir, M.H., Dewan, M.A.A., Chae, O.:
A dynamic histogram equalization for image contrast enhancement.
\emph{IEEE Transactions on Consumer Electronics} \textbf{53}(2), 593--600 (2007)

\bibitem[Arora and Anand(2018)]{arora2018novel}
Arora, S., Anand, P.:
A novel adaptive butterfly optimization algorithm.
\emph{International Journal of Computational Mathematics and Sciences} \textbf{7}(5), 26--34 (2018)

\bibitem[Arora and Singh(2019a)]{arora2019butterfly}
Arora, S., Singh, S.:
Butterfly optimization algorithm: a novel approach for global optimization.
\emph{Soft Computing} \textbf{23}(3), 715--734 (2019)



\bibitem{arora2019improved}
Arora, S., Singh, S.:
An improved butterfly optimization algorithm with chaos.
\emph{Journal of Intelligent \& Fuzzy Systems} \textbf{36}(6), 5879--5899 (2019)

\bibitem[Bhandari et~al.(2015)]{bhandari2015modified}
Bhandari, A.K., Kumar, A., Singh, G.K.:
Modified artificial bee colony based computationally efficient multilevel thresholding for satellite image segmentation using Kapur's, Otsu and Tsallis function.
\emph{Expert Systems with Applications} \textbf{42}(3), 1573--1601 (2015)

\bibitem[Buades et~al.(2005)]{buades2005non}
Buades, A., Coll, B., Morel, J.M.:
A non-local algorithm for image denoising.
In: \emph{IEEE Conference on Computer Vision and Pattern Recognition}, vol.~2, pp. 60--65 (2005)

\bibitem[Chen et~al.(2018)]{chen2018learning}
Chen, C., Chen, Q., Xu, J., Koltun, V.:
Learning to see in the dark.
In: \emph{IEEE Conference on Computer Vision and Pattern Recognition}, pp. 3291--3300 (2018)

\bibitem[Cui et~al.(2022)]{cui2022illumination}
Cui, Z., Li, K., Gu, L., Su, S., Gao, P., Jiang, Z., Qiao, Y., Haber, T.:
You only need 90K parameters to adapt light: a light weight transformer for image enhancement and exposure correction.
In: \emph{British Machine Vision Conference (BMVC)}, pp. 1--14 (2022)

\bibitem[Dabov et~al.(2007)]{dabov2007image}
Dabov, K., Foi, A., Katkovnik, V., Egiazarian, K.:
Image denoising by sparse 3-D transform-domain collaborative filtering.
\emph{IEEE Transactions on Image Processing} \textbf{16}(8), 2080--2095 (2007)

\bibitem[Deng et~al.(2022)]{deng2022low}
Deng, L., et~al.:
Low-illumination image enhancement algorithm based on improved multi-scale retinex and ABC algorithm optimization.
\emph{Frontiers in Bioengineering and Biotechnology} \textbf{10}, 865820 (2022)

\bibitem[Fan et~al.(2025)]{fan2025iniretinex}
Fan, G., Yao, Z., Chen, G.Y., Su, J.N., Gan, M.:
IniRetinex: rethinking retinex-type low-light image enhancer via initialization perspective.
In: \emph{AAAI Conference on Artificial Intelligence}, pp. 1--9 (2025)

\bibitem[Fister et~al.(2013)]{fister2013modified}
Fister, I., Yang, X.S., Brest, J., Fister~Jr, I.:
Modified firefly algorithm using quaternion representation.
\emph{Expert Systems with Applications} \textbf{40}(18), 7220--7230 (2013)

\bibitem[Gonzalez and Woods(2018)]{gonzalez2018digital}
Gonzalez, R.C., Woods, R.E.:
\emph{Digital Image Processing}, 4th edn.
Pearson, New York (2018)

\bibitem[Guo et~al.(2020)]{guo2020zero}
Guo, C., Li, C., Guo, J., et~al.:
Zero-reference deep curve estimation for low-light image enhancement.
In: \emph{IEEE/CVF Conference on Computer Vision and Pattern Recognition}, pp. 1780--1789 (2020)

\bibitem[He et~al.(2013)]{he2013guided}
He, K., Sun, J., Tang, X.:
Guided image filtering.
\emph{IEEE Transactions on Pattern Analysis and Machine Intelligence} \textbf{35}(6), 1397--1409 (2013)

\bibitem[Horng(2012)]{horng2012vector}
Horng, M.H.:
Vector quantization using the firefly algorithm for image compression.
\emph{Expert Systems with Applications} \textbf{39}(1), 1078--1091 (2012)

\bibitem[Huang et~al.(2013)]{huang2013efficient}
Huang, S.C., Cheng, F.C., Chiu, Y.S.:
Efficient contrast enhancement using adaptive gamma correction with weighting distribution.
\emph{IEEE Transactions on Image Processing} \textbf{22}(3), 1032--1041 (2013)

\bibitem[Ibrahim and Kong(2007)]{ibrahim2007brightness}
Ibrahim, H., Kong, N.S.P.:
Brightness preserving dynamic histogram equalization for image contrast enhancement.
\emph{IEEE Transactions on Consumer Electronics} \textbf{53}(4), 1752--1758 (2007)

\bibitem[Jiang et~al.(2021)]{jiang2021enlightengan}
Jiang, Y., Gong, X., Liu, D., et~al.:
EnlightenGAN: deep light enhancement without paired supervision.
\emph{IEEE Transactions on Image Processing} \textbf{30}, 2340--2349 (2021)

\bibitem[Jobson et~al.(1997a)]{jobson1997multiscale}
Jobson, D.J., Rahman, Z., Woodell, G.A.:
A multiscale retinex for bridging the gap between color images and the human observation of scenes.
\emph{IEEE Transactions on Image Processing} \textbf{6}(7), 965--976 (1997)



\bibitem{jobson1997properties}
Jobson, D.J., Rahman, Z., Woodell, G.A.:
Properties and performance of a center/surround retinex.
\emph{IEEE Transactions on Image Processing} \textbf{6}(3), 451--462 (1997)

\bibitem[Jyothi et~al.(2026)]{jyothi2026robust}
Jyothi, P., Haritha, D., Arava, K.:
A robust deep learning framework for accurate pose and gender prediction.
\emph{Multimedia Tools and Applications} \textbf{85}, 269 (2026)

\bibitem[Kennedy and Eberhart(1995)]{kennedy1995particle}
Kennedy, J., Eberhart, R.:
Particle swarm optimization.
In: \emph{IEEE International Conference on Neural Networks}, vol.~4, pp. 1942--1948 (1995)

\bibitem[Kim(1997)]{kim1997contrast}
Kim, Y.T.:
Contrast enhancement using brightness preserving bi-histogram equalization.
\emph{IEEE Transactions on Consumer Electronics} \textbf{43}(1), 1--8 (1997)

\bibitem[Land(1977)]{land1977retinex}
Land, E.H.:
The retinex theory of color vision.
\emph{Scientific American} \textbf{237}(6), 108--129 (1977)

\bibitem[Land and McCann(1971)]{land1971lightness}
Land, E.H., McCann, J.J.:
Lightness and Retinex theory.
\emph{Journal of the Optical Society of America} \textbf{61}(1), 1--11 (1971)

\bibitem[Li et~al.(2020)]{li2020enhanced}
Li, H., et~al.:
Enhanced retinex algorithm for mine image enhancement using improved bootstrap filter.
\emph{Journal of China Coal Society} \textbf{45}(S2), 1010--1018 (2020)

\bibitem[Liu et~al.(2021a)]{liu2021benchmarking}
Liu, J., Xu, D., Yang, W., Fan, M.:
Benchmarking low-light image enhancement and beyond.
\emph{International Journal of Computer Vision} \textbf{129}, 1153--1184 (2021)

\bibitem[Liu et~al.(2021b)]{liu2021retinex}
Liu, R., Ma, L., Zhang, J., Fan, X., Luo, Z.:
Retinex-inspired unrolling with cooperative prior architecture search for low-light image enhancement.
In: \emph{IEEE/CVF Conference on Computer Vision and Pattern Recognition}, pp. 10561--10570 (2021)

\bibitem[Loh and Chan(2019)]{loh2019getting}
Loh, Y.P., Chan, C.S.:
Getting to know low-light images with the exclusively dark dataset.
\emph{Computer Vision and Image Understanding} \textbf{178}, 30--42 (2019)

\bibitem[Ma et~al.(2017)]{ma2017multiscale}
Ma, J., et~al.:
Multi-scale retinex with color restoration image enhancement based on Gaussian filtering and guided filtering.
\emph{International Journal of Modern Physics B} \textbf{31}(16), 1744077 (2017)

\bibitem[Ma et~al.(2022)]{ma2022toward}
Ma, L., Ma, T., Liu, R., Fan, X., Luo, Z.:
Toward fast, flexible, and robust low-light image enhancement.
In: \emph{IEEE/CVF Conference on Computer Vision and Pattern Recognition}, pp. 5637--5646 (2022)

\bibitem[Makhadmeh et~al.(2023)]{makhadmeh2023recent}
Makhadmeh, S.N., Al-Betar, M.A., Abasi, A.K., et~al.:
Recent advances in butterfly optimization algorithm, its versions and applications.
\emph{Archives of Computational Methods in Engineering} \textbf{30}, 1399--1420 (2023)

\bibitem[Mantegna(1994)]{mantegna1994fast}
Mantegna, R.N.:
Fast, accurate algorithm for numerical simulation of L\'{e}vy stable stochastic processes.
\emph{Physical Review E} \textbf{49}(5), 4677--4683 (1994)

\bibitem[Mirjalili(2019)]{mirjalili2019genetic}
Mirjalili, S.:
Genetic algorithm.
In: \emph{Evolutionary Algorithms and Neural Networks}, SCI, vol.~780, pp. 43--55. Springer (2019)

\bibitem[Mittal et~al.(2012)]{mittal2012no}
Mittal, A., Moorthy, A.K., Bovik, A.C.:
No-reference image quality assessment in the spatial domain.
\emph{IEEE Transactions on Image Processing} \textbf{21}(12), 4695--4708 (2012)

\bibitem[Mittal et~al.(2013)]{mittal2013making}
Mittal, A., Soundararajan, R., Bovik, A.C.:
Making a ``completely blind'' image quality analyzer.
\emph{IEEE Signal Processing Letters} \textbf{20}(3), 209--212 (2013)

\bibitem[Ning et~al.(2023)]{ning2023low}
Ning, X., Su, J., Kochan, O.:
Low-illumination image enhancement method based on adaptive MSRCR algorithm.
In: \emph{CEUR Workshop Proceedings}, vol.~3387, pp. 78--89 (2023)

\bibitem[Parthasarathy and Sankaran(2012)]{parthasarathy2012automated}
Parthasarathy, S., Sankaran, P.:
An automated multi scale retinex with color restoration for image enhancement.
In: \emph{IEEE National Conference on Communications}, pp. 1--5 (2012)

\bibitem[Petro et~al.(2014)]{petro2014multiscale}
Petro, A.B., Sbert, C., Morel, J.M.:
Multiscale retinex.
\emph{Image Processing On Line} \textbf{4}, 71--88 (2014)

\bibitem[Provenzi et~al.(2005)]{provenzi2005mathematical}
Provenzi, E., De~Carli, L., Rizzi, A., Marini, D.:
Mathematical definition and analysis of the retinex algorithm.
\emph{Journal of the Optical Society of America A} \textbf{22}(12), 2613--2621 (2005)

\bibitem[Sadia et~al.(2019)]{sadia2019color}
Sadia, H., Azeem, F., Ullah, H., et~al.:
Color image enhancement using multiscale retinex with guided filter.
\emph{Multimedia Tools and Applications} \textbf{78}, 12455--12475 (2019)

\bibitem[Sharma and Pant(2022)]{sharma2022butterfly}
Sharma, T.K., Pant, M.:
\emph{Butterfly Optimization Algorithm: Theory and Engineering Applications}.
Springer, Singapore (2022)

\bibitem[Su et~al.(2020)]{su2020multiscale}
Su, Z., et~al.:
Multi-scale gradient domain-guided filtering for low-illumination scene image enhancement.
\emph{IEEE Access} \textbf{8}, 176502--176515 (2020)

\bibitem[Talbi(2002)]{talbi2002taxonomy}
Talbi, E.G.:
A taxonomy of hybrid metaheuristics.
\emph{Journal of Heuristics} \textbf{8}(5), 541--564 (2002)

\bibitem[Tilahun and Ong(2012)]{tilahun2012modified}
Tilahun, S.L., Ong, H.C.:
Modified firefly algorithm.
\emph{Journal of Applied Mathematics} \textbf{2012}, 467631 (2012)

\bibitem[Wang et~al.(2004)]{wang2004image}
Wang, Z., Bovik, A.C., Sheikh, H.R., Simoncelli, E.P.:
Image quality assessment: from error visibility to structural similarity.
\emph{IEEE Transactions on Image Processing} \textbf{13}(4), 600--612 (2004)

\bibitem[Wang et~al.(2013)]{wang2013naturalness}
Wang, S., Zheng, J., Hu, H.M., Li, B.:
Naturalness preserved enhancement algorithm for non-uniform illumination images.
\emph{IEEE Transactions on Image Processing} \textbf{22}(9), 3538--3548 (2013)

\bibitem[Wang et~al.(2023)]{wang2023ultra}
Wang, T., Zhang, K., Shen, T., Luo, W., Stenger, B., Lu, T.:
Ultra-high-definition low-light image enhancement: a benchmark and transformer-based method.
In: \emph{AAAI Conference on Artificial Intelligence}, vol.~37, pp. 2654--2662 (2023)

\bibitem[Wei et~al.(2018)]{wei2018deep}
Wei, C., Wang, W., Yang, W., Liu, J.:
Deep retinex decomposition for low-light enhancement.
In: \emph{British Machine Vision Conference (BMVC)}, pp. 1--12 (2018)

\bibitem[Xu et~al.(2023)]{xu2023lowlight}
Xu, X., Wang, R., Lu, J.:
Low-light image enhancement via structure modeling and guidance.
In: \emph{IEEE/CVF Conference on Computer Vision and Pattern Recognition}, pp. 9893--9903 (2023)

\bibitem[Yan et~al.(2026)]{yan2026image}
Yan, B.J., Dong, H., Liu, Y.L.:
Image enhancement of foreign objects in underground coal transportation belt based on improved retinex algorithm.
\emph{Multimedia Tools and Applications} \textbf{85}, 271 (2026)

\bibitem[Yang(2009)]{yang2009firefly}
Yang, X.S.:
Firefly algorithms for multimodal optimization.
In: \emph{Stochastic Algorithms: Foundations and Applications}, LNCS, vol.~5792, pp. 169--178. Springer (2009)

\bibitem[Yin et~al.(2025)]{yin2025mambadpf}
Yin, S., et~al.:
MambaDPF-Net: a dual-path fusion network with selective state space modeling for robust low-light image enhancement.
\emph{Electronics} \textbf{14}(22), 4533 (2025)

\bibitem[Zamir et~al.(2020)]{zamir2020learning}
Zamir, S.W., Arora, A., Khan, S., et~al.:
Learning enriched features for real image restoration and enhancement.
In: \emph{European Conference on Computer Vision}, pp. 492--511 (2020)

\bibitem[Zhang et~al.(2019)]{zhang2019kindling}
Zhang, Y., Zhang, J., Guo, X.:
Kindling the darkness: a practical low-light image enhancer.
In: \emph{ACM International Conference on Multimedia}, pp. 1632--1640 (2019)

\bibitem[Zhang et~al.(2021)]{zhang2021improved}
Zhang, H., et~al.:
Improved bilateral filtering and multi-scale Retinex for image enhancement.
\emph{Journal of Image and Graphics} \textbf{26}(4), 681--693 (2021)

\bibitem[Zhi et~al.(2022)]{zhi2022enhanced}
Zhi, X., et~al.:
Enhanced Retinex algorithm combined with CLAHE for image contrast improvement.
\emph{Optik} \textbf{260}, 169027 (2022)

\bibitem[Zuiderveld(1994)]{zuiderveld1994contrast}
Zuiderveld, K.:
Contrast limited adaptive histogram equalization.
In: \emph{Graphics Gems IV}, pp. 474--485. Academic Press (1994)

\end{thebibliography}
\end{document}